\documentclass[journal]{IEEEtran}
\usepackage{hyperref}
\usepackage{graphicx}
\usepackage{caption}
\usepackage{subcaption}
\usepackage{enumitem}
\usepackage{amsfonts}
\usepackage{bbm}
\usepackage{footnote}
\usepackage{amsmath}
\usepackage{enumitem}
\usepackage{xcolor}
\usepackage{placeins}
\usepackage{fancyhdr}

\usepackage{xspace}
\makeatletter
\DeclareRobustCommand\onedot{\futurelet\@let@token\@onedot}
\def\@onedot{\ifx\@let@token.\else.\null\fi\xspace}
\def\eg{\emph{e.g}\onedot} 
\def\ie{\emph{i.e}\onedot}

\def\etal{\emph{et al}\onedot}
\makeatother

\usepackage{tikz}
\usepackage{lipsum}

\begin{document}
\title{Perceiving Humans: \\ from Monocular 3D Localization to \\ Social Distancing}
\author{Lorenzo~Bertoni,
        Sven~Kreiss,
        Alexandre~Alahi
\thanks{Lorenzo Bertoni, Sven Kreiss and Alexandre Alahi are at the VITA lab at EPFL,
1015~Lausanne, Switzerland,
e-mail: lorenzo.bertoni@epfl.ch}}

\maketitle

\begin{abstract}
Perceiving humans in the context of Intelligent Transportation Systems (ITS) often relies on multiple cameras or expensive LiDAR sensors. In this work, we present a new cost-effective vision-based method that perceives humans' locations in 3D and their body orientation from a single image. We address the challenges related to the ill-posed monocular 3D tasks by proposing a neural network architecture that predicts confidence intervals in contrast to point estimates. Our neural network estimates human 3D body locations and their orientation with a measure of uncertainty. Our proposed solution (i) is privacy-safe, (ii) works with any fixed or moving cameras, and (iii) does not rely on ground plane estimation. We demonstrate the performance of our method with respect to three applications: locating humans in 3D, detecting social interactions, and verifying the compliance of recent safety measures due to the COVID-19 outbreak. We show that it is possible to rethink the concept of ``social distancing" as a form of social interaction in contrast to a simple location-based rule. We publicly share the source code towards an open science mission.

\end{abstract}

\IEEEpeerreviewmaketitle

\section{Introduction}


Over the past decades, we have witnessed new emerging technologies to localize humans in 3D, ranging from vision-based \cite{ llorca2012stereo, alahi2011sparsity, palffy2019occlusion, arnold2019survey, monoloco}, to LiDAR-based solutions \cite{liu2020tanet, hotspotnet} and multi-sensor approaches \cite{liang2018deep, xu2018pointfusion}. 
On one hand, vision-based technologies can capture detailed body poses and texture properties, but rely on a costly calibrated network of cameras \cite{alahi2014robust, delannay2009detection, hu20073d}. On the other hand, LiDAR sensors are limited by high cost, noise in case of adverse weather, and sparsity of point clouds over long ranges \cite{zhou2018voxelnet, qi2018frustum, arnold2019survey}. In this work, we show that given a single cost-effective RGB camera, we can not only extract humans' 3D locations but also their body orientations. Consequently, we can go beyond monocular 3D localization of humans and 
detect social interactions (\textit{e.g.}, whether two people are talking to each other) in transportation hubs, and even verify compliance with the recent safety measures due to the COVID-19 outbreak.

The COVID-19 pandemic has forced authorities to limit non-essential movements of people, especially in crowded areas or public transport \cite{cvd19worldbank}. Social distancing measures are becoming essential to restart passenger services, \eg, leaving train seats unoccupied. Yet in many contexts it is not obvious how to preserve inter-personal distances.
When the risk of contagion remains, we should work to minimize it, and perceiving social interactions can play a vital role. In fact, talking with a person does not incur the same risk of infection as passing someone in the street. The infection rate of a disease can be summarized as the product of exposure time and exposure to virus particles \cite{remington1985airborne, cvd19speaking}. When people are talking together, not only does the exposure time escalate, but the act of speaking itself increases the release of respiratory droplets about ten fold \cite{asadi2019aerosol, cvd19airborne}.
These analyses urge us to rethink safety measures and focus on proximal social interactions, which can be defined as any behavior of two or more people mutually oriented towards each other and who influence or take into account each other's subjective experiences or intentions \cite{rummel1981understanding}.  
We show that we can monitor the concept of ``social distancing" as a form of social interaction in contrast to a simple location-based rule or smartphone-based solutions \cite{zhao2000mobile, zandbergen2009accuracy, kasemsuppakorn2013pedestrian}. A few methods have studied interactions from images \cite{cristani2011social, cristani2011towards}, but their results are either limited to personal photos, \cite{yang2012recognizing}, indoor scenarios, \cite{aimar2019social}, or necessitate a homography calibration \cite{cristani2011social, cristani2011towards}. However,
the study of social distancing
requires an understanding of social interactions in a variety of unconstrained scenarios, either outdoors or within large facilities. 

 \begin{figure*}
  \centering
  \includegraphics[width=0.99\linewidth]{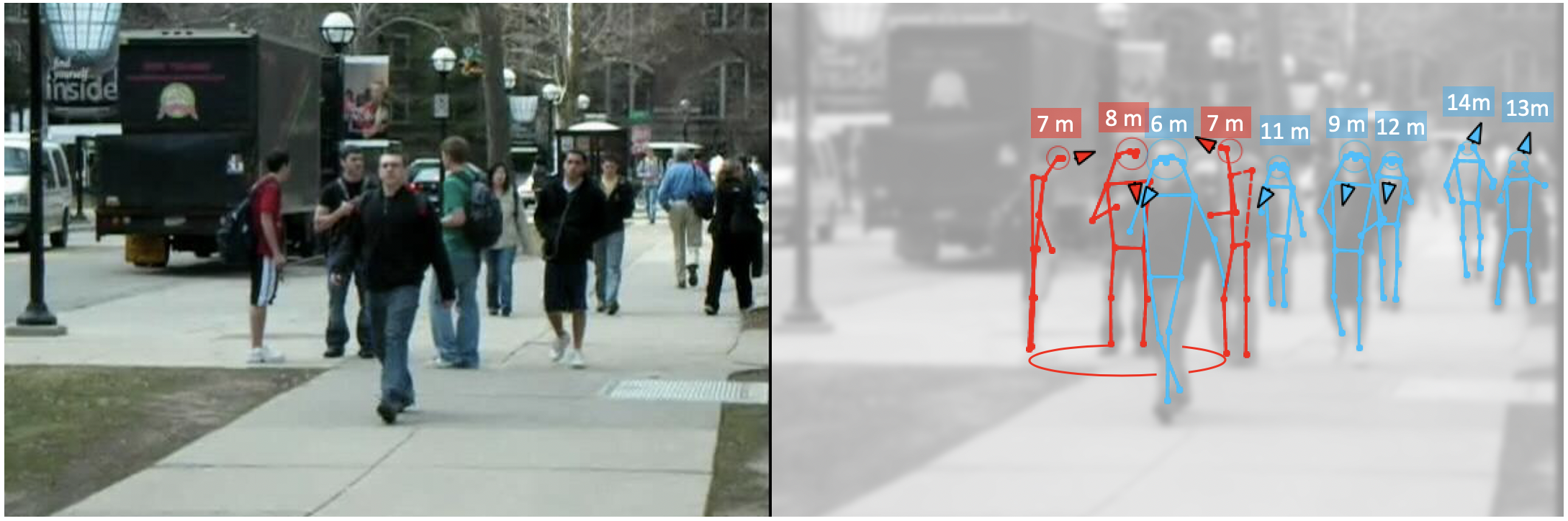}
  \caption{Our method retrieves 3D locations with confidence intervals, body orientations, social interactions and social distancing in the wild from a single RGB image. We leverage 2D human poses as intermediate representations to verify social distancing compliance while preserving privacy.}
  \label{fig:pull}
\end{figure*}

In this paper, we propose a deep learning approach that perceives humans and their social interactions in the 3D space from visual cues only. We argue that the fundamental challenge behind recognizing social interactions from a monocular camera is to perceive humans in 3D, an intrinsically ill-posed problem. We address this ambiguity by predicting confidence intervals in contrast to point estimates through a loss function based on the Laplace distribution. Our approach consists of three main steps.
First, we use an off-the-shelf pose detector \cite{kreiss2019pifpaf} to obtain 2D keypoints, a low-dimensional representation of humans. Second, the 2D poses are fed into a light-weight feed-forward neural network that predicts 3D locations, orientations and corresponding confidence intervals for each person. Finally, driven by these perception tasks, we aim at investigating how people use the space when interacting in groups. According to the subfield of proxemics, people tend to arrange themselves spontaneously in specific configurations called F-formations \cite{kendon1990conducting}. The detection of F-formations is critical to infer social relations \cite{cristani2011social, cristani2011towards}. Our intuition is that knowing the 3D location and orientation of people in a scene allows the accurate retrieval of F-formations with simple probabilistic rules. Inspired by \cite{cristani2011social, cristani2011towards}, we exploit our predicted confidence intervals to develop a simple probabilistic approach to detect F-formations and social interactions among humans. Consequently, we show that we can redefine the concept of social distancing to go beyond a simple measure of distance. We provide simple rules to verify safety compliance in indoor/outdoor scenarios based on the interactions among people rather than their relative position alone. Finally, the design of our pipeline encourages privacy-safe implementations by decoupling the image processing step. Our network is trained on and performs inference with anonymous 2D human poses. An example is provided in Figure \ref{fig:pull}, where 3D location, orientation and interactions among people are analyzed to verify social distancing compliance in a private manner.

Technically, our main contributions are three-fold:
(i) we outperform monocular methods for the 3D localization task on the publicly-available KITTI dataset \cite{Geiger2013Kitti} while also estimating meaningful confidence intervals; (ii) we effectively capture social interactions among people on the Collective Activity Dataset \cite{choi2009they} without any additional training or homography estimation; (iii) we show that we can redefine the concept of social distancing based on social cues while preserving the privacy of its users. Our code is publicly available\footnote{\url{https://github.com/vita-epfl/monoloco}}.

\section{Related Work}
In this work, we tackle the high-level task of understanding 3D spatial relations among humans from a single RGB image without ground plane estimation. The core of our pipeline is composed of a sequence of low-level tasks to process the image and extract 3D information, which can be called monocular 3D vision. The more general field of computer vision has experienced a fundamental transition towards data-hungry deep learning methods thanks to their natural ability to process data in raw form \cite{lecun2015deep}. The transition started with 2D tasks, such as object detection \cite{ren2015faster, yolo} and human pose estimation \cite{openpose}, and it expanded to 3D tasks such as 3D object detection \cite{m3d, wenlongPSF, monstereo}, object recognition \cite{muzahid2020curvenet}, depth estimation \cite{godard2017monodepth}, or even forecasting tasks \cite{alahi2017learning}. A crucial factor in this transformation has been the release of massive datasets for 2D \cite{deng2009imagenet, Lin2014MicrosoftCC, sun2020proximity} and 3D tasks \cite{Geiger2013Kitti, nuscenes, argoverse, waymo,kothari_human_2020}, especially in the context of autonomous driving. While perception tasks have been monopolized by relatively new deep learning algorithms, the study of social interactions is based on historic discoveries in behavioural science.  In this work, we only focus on \textit{proxemics}: the subfield relating human interactions with the use of space \cite{hall1966hidden}. The remainder of this section is organized as follows. First, we
review 2D and 3D  tasks that compose our perception pipeline, namely human pose estimation, monocular 3D object detection, and uncertainty estimation. Last, we focus on the study of proxemics and its applications for computer vision and transportation research.

\subsection{Monocular 3D Vision}
We include three different sub tasks under the ``Monocular 3D Vision" umbrella, as they all contribute to perceive humans in the 3D space from single RGB images. We are interested in algorithms that can operate in outdoor and crowded environments, so when applicable, we focus our review on perception techniques for autonomous driving.

\vspace{5pt}
\textbf{Human Pose Estimation. }
Detecting people in images and estimating their skeleton is a widely studied problem. 
State-of-the-art methods are based on Convolutional Neural Networks and can be grouped into top-down  and bottom-up methods. 
Top-down approaches consist in detecting each instance in the image first and then estimating body joints within the boundaries of the inferred bounding box \cite{Papandreou2017TowardsAM, Fang2017RMPERM, He2017MaskR, xiao2018simple}. 
Bottom-up approaches estimate separately each body joint through convolutional architectures and then combine them to obtain a full human pose \cite{openpose,Cao2017RealtimeM2, newell2017associative, personlab, kocabas2018multiposenet}. More recently PifPaf \cite{kreiss2019pifpaf, kreiss2021openpifpaf} proposed a method tailored for autonomous driving scenarios that performs well in low-resolution, crowded and occluded scenes. Related to our work is Simple Baseline~\cite{martinez2017simple}, which shows the effectiveness of latent information contained in 2D joints stimuli. They achieve state-of-the-art results by simply predicting 3D joints from 2D poses through a light, fully connected network. However, these lines of work estimate relative 3D joint positions \cite{MorenoNoguer20173DHP, zanfir2018deep, Rogez2019LCRNetM2}, or relative 3D meshes \cite{smpl2015,kanazawa2018end}, not providing any information about the real 3D location in the scene.

\vspace{5pt}
\textbf{Monocular 3D Object Detection.} The majority of approaches for monocular 3D object detection in the transportation domain focus on vehicles as they are rigid objects with known shape. Very recently, a few works have extended their approaches to the pedestrian category. MonoPSR \cite{ku2019monopsr} evaluates pedestrians from monocular RGB images, leveraging point clouds at training time to learn local shapes of objects. MonoDIS \cite{monodis-tpami} proposes to disentangle the contribution of each loss component, while SMOKE \cite{smoke20} combines a single keypoint estimate with regressed 3D variables.
Kundegorski and Breckon \cite{Kundegorski2014APA} achieve reasonable performances combining infrared imagery and real-time photogrammetry. Alahi \etal combine monocular images with wireless signals \cite{alahi2015rgb} or with additional visual priors \cite{alahi2008object,alahi2014robust,alahi2017tracking}. 
The seminal work of Mono3D \cite{m3d} exploits deep learning to create 3D object proposals for \textit{car}, \textit{pedestrian} 
and \textit{cyclist} categories but it does not evaluate 3D localization of pedestrians. 
It assumes a fixed ground plane orthogonal to the camera and the proposals are then scored based on scene priors, such as shape, 
semantic and instance segmentations. 
The following methods continue to leverage Convolutional Neural Networks and focus only on \textit{Car} instances. 
To regress 3D pose parameters from 2D detections, Deep3DBox \cite{mousavian20173d}, MonoGRnet \cite{qin2019monogrnet}, and Hu \etal \cite{Hu2018JointM3} use geometrical reasoning for 3D localization, 
while Multi-fusion \cite{xu2018multi} and ROI-10D \cite{manhardt2019roi} incorporate a module for depth estimation. 
Roddick \etal \cite{roddick2018orthographic} escape the image domain by mapping image-based features into a birds-eye view representation using integral images. 
Another line of work fits 3D templates of cars to the image~\cite{Xiang2015Datadriven3V, xiang2017subcategory, Chabot2017DeepMA, 3d-rcnn}.
While many of the related methods achieve reasonable performances for vehicles, current literature lacks monocular methods addressing other categories in the context of autonomous driving, such as pedestrians and cyclists.

\vspace{5pt}
\textbf{Uncertainty Estimation in Computer Vision.}
Deep neural networks need the ability not only to provide the correct outputs but also a measure of uncertainty, 
especially in safety-critical scenarios like autonomous driving. 
Traditionally, Bayesian Neural Networks~\cite{Richard1991NeuralNC, neal2012bayesian} are used to model epistemic uncertainty 
through probability distributions over the model parameters. However, these distributions are often intractable 
and researchers have proposed interesting solutions to perform approximate Bayesian inference to measure uncertainty,
including Variational Inference \cite{graves2011practical, blundell15weight, salimans2015markov} 
and Deep Ensembles \cite{lakshminarayanan2017simple}. Alternatively, Gal \etal \cite{Gal2016Dropout, gal2017concrete} show that 
applying dropout \cite{srivastava2014dropout} at inference time yields a form of variational inference where a mixture of multivariate Gaussian distributions with small variances models the network parameters. 
This technique, called Monte Carlo (MC) dropout, has earned great popularity due to its adaptability to non-probabilistic deep learning frameworks. Very recently,  Postels \etal \cite{postels2019sampling} proposed a sampling-free method to approximate epistemic uncertainty, treating noise injected in a neural network as errors on the activation values.
In computer vision, uncertainty estimation using MC dropout has been applied for depth regression tasks 
\cite{Kendall2017WhatUD, postels2019sampling}, scene segmentation \cite{mukhoti2018evaluating, Kendall2017WhatUD} 
and, more recently, LiDAR 3D object detection for cars \cite{feng2018towards}. In this work, we demonstrate its relevance for monocular human 3D localization.

\subsection{Social Interactions}\label{sec:related_social}
We aim to capture social interactions among people and monitor social distancing from visual cues.
Related works include the broad field of behavioral science \cite{joo2019towards}. Here we focus on the subfield called \textit{proxemics}, which investigates how people use and organize the space they share with others \cite{hall1966hidden, cristani2011towards}. People tend to arrange themselves spontaneously in specific configurations called F-formations \cite{kendon1990conducting}. These formations are characterized by an internal empty zone (o-space) surrounded by a concentric ring where people are located (p-space). According to Kendon \cite{kendon1990conducting}: ``\textit{an F-formation arises whenever two or more people sustain a spatial and orientational relationship in which the space between them is one to which they have equal, direct, and exclusive access"}.

These formations characterize how people use the space when  interacting with each other. They are characterized by three types of social spaces \cite{hall1966hidden, cristani2011social}:
\begin{enumerate}
\item \textit{o-space}:  A circular empty region to preserve the personal space of the participants around it. Every participant looks inward and no people are allowed inside. The type of relation (\eg, personal or business-related) defines the dimensions of this space,
\item \textit{p-space}: a concentric ring around the o-space that contains all the participants, 
\item \textit{r-space}: the area outside the p-space.
\end{enumerate} 

In the case of two participants, typical F-formations are vis-a-vis, L-shape, and side-by-side. For larger groups, a circular formation is typically formed \cite{kuzuoka2010reconfiguring}. An example of an F-formation configuration is shown in Figure \ref{fig:o-space}.

To the best of our knowledge, Cristani  \etal 2011a \cite{cristani2011social} is the first work to focus solely on visual cues to discover F-formations and social interactions. In parallel, Cristani \etal 2011b \cite{cristani2011towards} study how people get closer to each other when the social relation is more intimate. The following works have proposed various techniques to automatically detect F-formations in heterogeneous real crowded scenarios \cite{tran2013social, bazzani2013social,vascon2014game,setti2015f}. In all approaches, it is clear how the detection of F-formations is critical to infer social relations and we decide to follow their lead. 
This line of work, however, considers as input the position of people on the ground floor and their orientation \cite{cristani2011towards} or requires a homography estimation to compute the x-y-z coordinates of humans \cite{cristani2011social}.  On the contrary, our approach works end-to-end from a single RGB image. The perception stage, \ie, extracting 3D detections from a monocular image, is arguably the most challenging one due to the intrinsic ambiguity of perspective projections.

Finally, social interactions have also been studied in the context of personal photos \cite{yang2012recognizing} or egocentic photo-streams \cite{aghaei2015towards,nakamura2017jointly,aimar2019social}. Both approaches assume humans to stand less than a few meters apart from each other and the camera, and do not scale to long range applications, such as monitoring an airport terminal.
Recently, deep learning approaches have been adopted to understand social interactions under a different perspective. Joo \etal \cite{joo2019towards} learn to predict behavioral cues of a target person (\eg, body orientation) from the position and orientation of another person. They learn the dynamics between social interactions in a data-driven manner, laying the foundations for deep learning to be applied in the field of behavioral science.

 \begin{figure*}
  \centering
  \includegraphics[width=\linewidth]{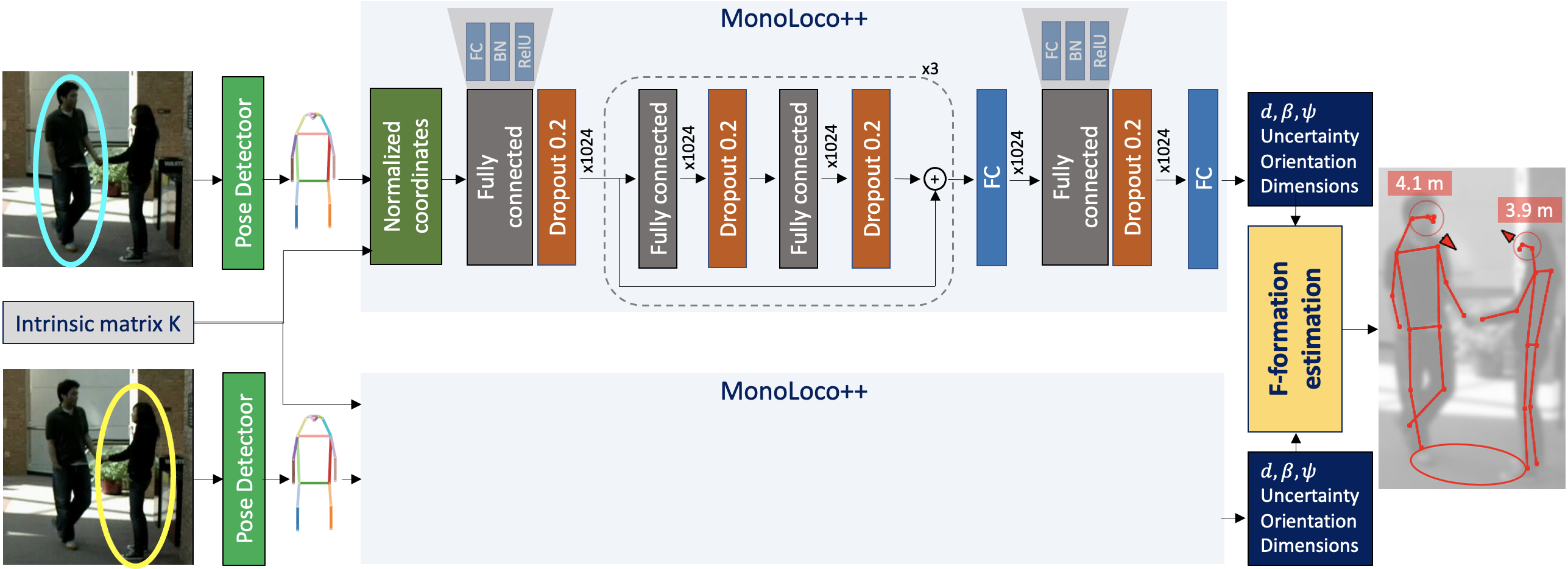}
  \caption{Overall architecture. \textbf{MonoLoco++}:
  the input is a set of 2D joints extracted from a raw image and the output is the 3D location, orientation and dimensions of a human with the localization uncertainty. 3D location is estimated with spherical coordinates: radial distance $d$, azimuthal angle $\beta$, and polar angle $\psi$. Every fully connected layer (FC) outputs 1024 features and is followed by a Batch Normalization layer (BN) \cite{ioffe2015batch} and a ReLU activation function. \textbf{F-formations}: all estimates from MonoLoco++ are analyzed with an \textit{all-vs-all} approach to discover F-formations and estimate social interactions/distancing using Eq.~\ref{eq:si}. }
 
  \label{fig:method}
\end{figure*}

\section{3D Localization Ambiguity}\label{sec:ambiguity}
A critical challenge in understanding social interactions from visual cues is the 3D localization pillar. Inferring distance of humans from monocular images is a fundamentally ill-posed problem. The majority of previous works have circumvented this challenge by assuming a planar ground plane and estimating a homography by manual measurement or by knowing some reference elements \cite{se2002ground,cristani2011social,m3d,3dop}. These approaches do not work when people are on stairs and they require a static calibrated setup. In this work, we address their limitations by directly estimating distance of humans without relying on a ground plane or homography.
This problem is ill-posed due to human variation of height. If every pedestrian has the same height, there would be no ambiguity. However, does this ambiguity prevent from robust localization? This section is dedicated to explore this question and analyze the maximum accuracy expected from monocular pedestrian localization. 

We are interested in the 3D localization error due to the ambiguity of perspective projection. Our approach consists in assuming that all humans have the same height $h_\textrm{mean}$ and analyzing the error of this assumption.
Inspired by Kundegorski and Breckon \cite{Kundegorski2014APA}, we model the localization error related to height variation as a function of the ground-truth distance from the camera, which we call \textit{task error}.
From the triangle similarity relation of human heights and distances, $ d_\textrm{h-mean} /h_\textrm{mean} = d_{gt} / h_{gt}$,
where $h_{gt}$ and $d_{gt}$ are the ground-truth human height and distance, $h_\textrm{mean}$ is the assumed mean height of a person and $d_\textrm{h-mean}$ the estimated distance under the $h_\textrm{mean}$ assumption.
We can define the task error for any person instance in the dataset as:
\begin{equation}
 e \equiv |d_{gt} - d_\textrm{h-mean}| = d_{gt} \; \left| 1 - \frac{h_\textrm{mean}}{h_{gt}} \right| \;\;\;.
\end{equation}
Previous studies from a population of 63,000 European adults have shown that the average height is $178 cm$ for males and $165 cm$ for females 
with a standard deviation of around $7 cm$ in both cases~\cite{visscher2008sizing}. However, a pose detector does not distinguish between genders. 
Assuming that the distribution of human stature follows a Gaussian distribution for male and female populations \cite{freeman1995cross}, 
we define the combined distribution of human heights, a Gaussian mixture distribution $P(H)$, as our unknown ground-truth height distribution. The \textit{expected task error} becomes
\begin{equation}
\hat{e} = d_{gt} \; E_{h \sim P(H) }\left[\left|1 - \frac{h_{mean}}{h}\right|\right] \;\;\; , 
 \label{eq:task_error}
\end{equation}
which represents a lower bound for monocular 3D pedestrian localization due to the intrinsic 
ambiguity of the task.
The analysis can be extended beyond adults. A 14-year old male reaches about~$90\%$ of his full height and a female about~$95\%$~\cite{freeman1995cross, Kundegorski2014APA}. 
Including people down to 14 years old leads to an additional source of height variation of~$7.9\%$ and~$5.6\%$ for men and women, respectively~\cite{Kundegorski2014APA}.
Figure~\ref{fig:results} shows the expected localization error $\hat{e}$ due to height variations in different cases as a linear function of the ground-truth distance from the camera~$d_{gt}$.  For a pedestrian 20 meters far, the localization error is approximately 1 meter.
This analysis shows that the ill-posed problem of localizing humans, while imposing an intrinsic limit, does not prevent from a good enough localization in many applications.

\section{Proposed Method}

The goals of our method are (i) to detect humans in 3D given a single image and (ii) to leverage this information to recognize social interactions and monitor social distancing. Figure \ref{fig:method} illustrates our overall method, which consists of three main steps. 
First, we exploit a pose detector to escape the image domain and reduce the input dimensionality. 2D human joints are a meaningful low-level representation which provides invariance to many factors, including background scenes, lighting, textures and clothes. Second, we use the 2D joints as input to a feed-forward neural network that predicts x-y-z coordinates and the associated uncertainty, orientation, and dimensions of each pedestrian. In the training phase, there is no supervision for the localization ambiguity. The network implicitly learns it from the data distribution. Third, the network estimates are combined to obtain F-formations \cite{hall1966hidden} and recognize social interactions.

\subsection{3D Human Detection}
The task of 3D object detection is defined as detecting 3D location of objects along with their orientation and dimensions \cite{Geiger2013Kitti, nuscenes}. The ambiguity of the task derives from the localization component as described in Section \ref{sec:ambiguity}.
Hence, we argue that effective monocular localization implies not only accurate estimates of the distance but also realistic predictions of uncertainty.
Consequently, we propose a method which learns the ambiguity from the data and predicts confidence intervals in contrast to point estimates. 
The task error modeled in Eq. \ref{eq:task_error} allows us to compare the predicted confidence intervals with the intrinsic ambiguity of the task.

\vspace{5pt}
\textbf{Input. } 
We use a pose estimator to detect a set of keypoints $\left[u_i, v_i \right]^T $ for every instance in the image. 
We then back-project each keypoint $i$ into normalized image coordinates $ \left[x^*_i, y^*_i, 1 \right]^T $ using 
the camera intrinsic matrix K:

\begin{equation}
\left[x^*_i, y^*_i, 1 \right]^T = K^{-1} \left[u_i, v_i, 1 \right]^T.
\label{eq:k}
\end{equation}
This transformation is essential to prevent the method from overfitting to a specific camera.

\vspace{5pt}
\textbf{2D Human Poses. }
We obtain 2D joint locations of humans using the off-the-shelf pose detector PifPaf~\cite{kreiss2019pifpaf, kreiss2021openpifpaf}, a state-of-the-art, bottom-up method designed for crowded scenes and occlusions.
The detector can be regarded as a stand-alone module independent from our network, which uses 2D joints as inputs. 
PifPaf has not been fine-tuned on any additional dataset for 3D object detection as no annotations for 2D poses are available.

\vspace{5pt}
\textbf{Output. }
We predict 3D localization, dimensions, and viewpoint angle with a regressive model. Estimating depth is arguably the most critical component in vision-based 3D object detection due to intrinsic limitations of monocular settings described in Section \ref{sec:ambiguity}.  However, due to perspective projections, an error in depth estimation $z$ would also affect the horizontal and vertical components $x$ and $y$. To disentangle the depth ambiguity from the other components, we use a spherical coordinate system  $(d, \beta, \psi)$, namely radial distance $d$, azimuthal angle $\beta$, and polar angle $\psi$. Another advantage of using a spherical coordinate system is that the size of an object projected onto the image plane directly depends on its radial distance $d$ and not on its depth $z$ \cite{monoloco}.
The same pedestrian in front of a camera or at the margin of the camera field-of-view will appear as having the same height 
in the image plane, as long as the distance from the camera $d$ is the same.

As already noted in \cite{li2019stereo}, the viewpoint angle is not equal to the object orientation as people at different locations may share the same orientation $\theta$ but results in different projections. Hence, we predict the viewpoint angle $\alpha$, which is defined as $\alpha=\theta + \beta$, where $\beta$ denotes the azimuth of the pedestrian with respect to the camera. Similarly to \cite{li2019stereo}, we also parametrize the angle as $\left[\sin \alpha, \cos \alpha \right] $ to avoid discontinuity. 
Regarding bounding box dimensions, we follow the standard procedure to calculate width, height and length of each pedestrian. We calculate average dimensions from the training set and regress the displacement from the expectation.

\vspace{5pt}
\textbf{Minimization Objective. }
Our final loss is the logarithm of the probability that all
components are ``well'' predicted, \ie, it is the sum of the log-probabilities
for the individual components. For every component but the 3D localization, we use a vanilla $L_1$ loss. To regress distances of people, we use a Laplace-based L1 loss \cite{Kendall2017WhatUD}, which we describe in the following section. Our minimization objective is a simple non-weighted sum of each loss function.


\vspace{5pt}
\textbf{Base Network. }
The building blocks of our model are shown in Figure \ref{fig:method}. 
The architecture, inspired by Martinez \etal \cite{martinez2017simple}, is a simple, deep, fully-connected network with six linear layers with 1024 output features. 
It includes dropout \cite{srivastava2014dropout} after every fully connected layer, 
batch-normalization \cite{ioffe2015batch} and residual connections \cite{he2016residual}. 
The model contains approximately 8M training parameters. 

\vspace{5pt}
\textbf{MonoLoco++ vs MonoLoco. }
We refer to our method as MonoLoco++. Technically, it differs from the previous MonoLoco \cite{monoloco} by:
\begin{itemize}
    \item the multi-task approach that combines 3D localization, orientation and bounding-box dimensions,
    \item the use of spherical coordinates to disentangle the ambiguity in the 3D localization task,
    \item an improved neural network architecture.
\end{itemize}
Combining precise 3D localization and orientation paves the way for activity recognition and social distancing, which was not possible using MonoLoco \cite{monoloco}. As illustrated in Fig \ref{fig:method}, multiple MonoLoco++ estimates are combined into the \textit{F-formation estimation} block to detect social interactions and social distancing. In addition, we will show how the above technical improvements benefit the monocular 3D localization task itself.

\subsection{Uncertainty}
In this work, we propose a probabilistic network which models two types of uncertainty: \textit{aleatoric} and \textit{epistemic} \cite{der2009aleatory, Kendall2017WhatUD}.
Aleatoric uncertainty is an intrinsic property of the task and the inputs. It does not decrease when collecting more data.
In the context of 3D monocular localization, the intrinsic ambiguity of the task represents
a quota of aleatoric uncertainty. In addition, some inputs may be more noisy than others, leading to an input-dependent aleatoric uncertainty.
Epistemic uncertainty is a property of the model parameters, and it can be reduced by gathering more data. It is useful to quantify the ignorance of the model about the collected data, \textit{e.g.}, in case of out-of-distribution samples.

\vspace{5pt}
\textbf{Aleatoric Uncertainty. }
Aleatoric uncertainty is captured through a probability distribution over the model outputs.
We define a relative Laplace loss based on the negative log-likelihood of a Laplace distribution as:
\begin{equation}
  L_{\textrm{Laplace}}(x|d,b) = \frac{|1-d/x|}{b} + \log(2b) \;\;\; ,
\label{eq:laplace} 
\end{equation}
where $x$ represents the ground-truth distance, $d$ the predicted distance, and $b$ the spread, making this training objective an attenuated $L_1$-type loss via spread $b$.

During training, the model has the freedom to ignore noisy data and attenuate its gradients by predicting a large spread $b$. As a consequence, inputs with high uncertainty have a small effect on the loss, making the network more robust to noisy data. The uncertainty is estimated in an unsupervised way, since no supervision is provided.
At inference time, the model predicts a Laplace distribution parameterized by the distance $d$ and a spread $b$. The latter one indicates the model's confidence about the predicted distance.
Following~\cite{Kendall2017WhatUD}, to avoid the singularity for $b=0$, 
we apply a change of variable to predict the log of the spread $s = \log(b)$.

Compared to previous methods \cite{Kendall2017WhatUD, wirges2019capturing}, 
we design a Laplace loss that works with relative distances to take into account the role of distance in our predictions. 
For example in autonomous driving scenarios, estimating the distance of a pedestrian with an absolute error can lead to a fatal accident if the person is very close, 
or be negligible if the same human is far away from the camera. 

\vspace{5pt}
\textbf{Epistemic Uncertainty. }
To model epistemic uncertainty, we follow \cite{Gal2016Dropout, Kendall2017WhatUD} and consider each parameter 
as a mixture of two multivariate Gaussians with small variances and means $0$ and $\theta$. 
The additional minimization objective for N data points is:
\begin{equation}
  L_{\textrm{dropout}}(\theta, p_{drop}) = \frac{1-p_{drop}}{2N}||\theta||^2 \;\;\; .
 \label{mc_drop}
\end{equation}

In practice, we perform dropout variational inference by training the model with dropout before every weight layer 
and then performing a series of stochastic forward passes at test time using the same dropout probability $p_{drop}$ of training time. 
The use of fully-connected layers makes the network particularly suitable for this approach, 
which does not require any substantial modification of the model.

The combined epistemic and aleatoric uncertainties are captured by the sample variance of predicted distances $\tilde{x}$. They are sampled from multiple Laplace distributions parameterized with the predictive distance $d$ and spread $b$ from multiple forward passes with MC dropout:
\begin{align}
  Var(\tilde{X}) =
  & \frac{1}{TI} \sum_{t=1}^T \sum_{i=1}^I \tilde{x}_{t,i}^2(d_t, b_t) - \left[ \frac{1}{TI} \sum_{t=1}^T \sum_{i=1}^I \tilde{x}_{t,i}(d_t, b_t) \right]^2 \;\;\; ,
\label{eq:variance}
\end{align}
where for each of the $T$ computationally expensive forward passes, $I$ computationally cheap samples are drawn from the Laplace distribution.

\begin{figure}
  \centering
  \includegraphics[width=\linewidth]{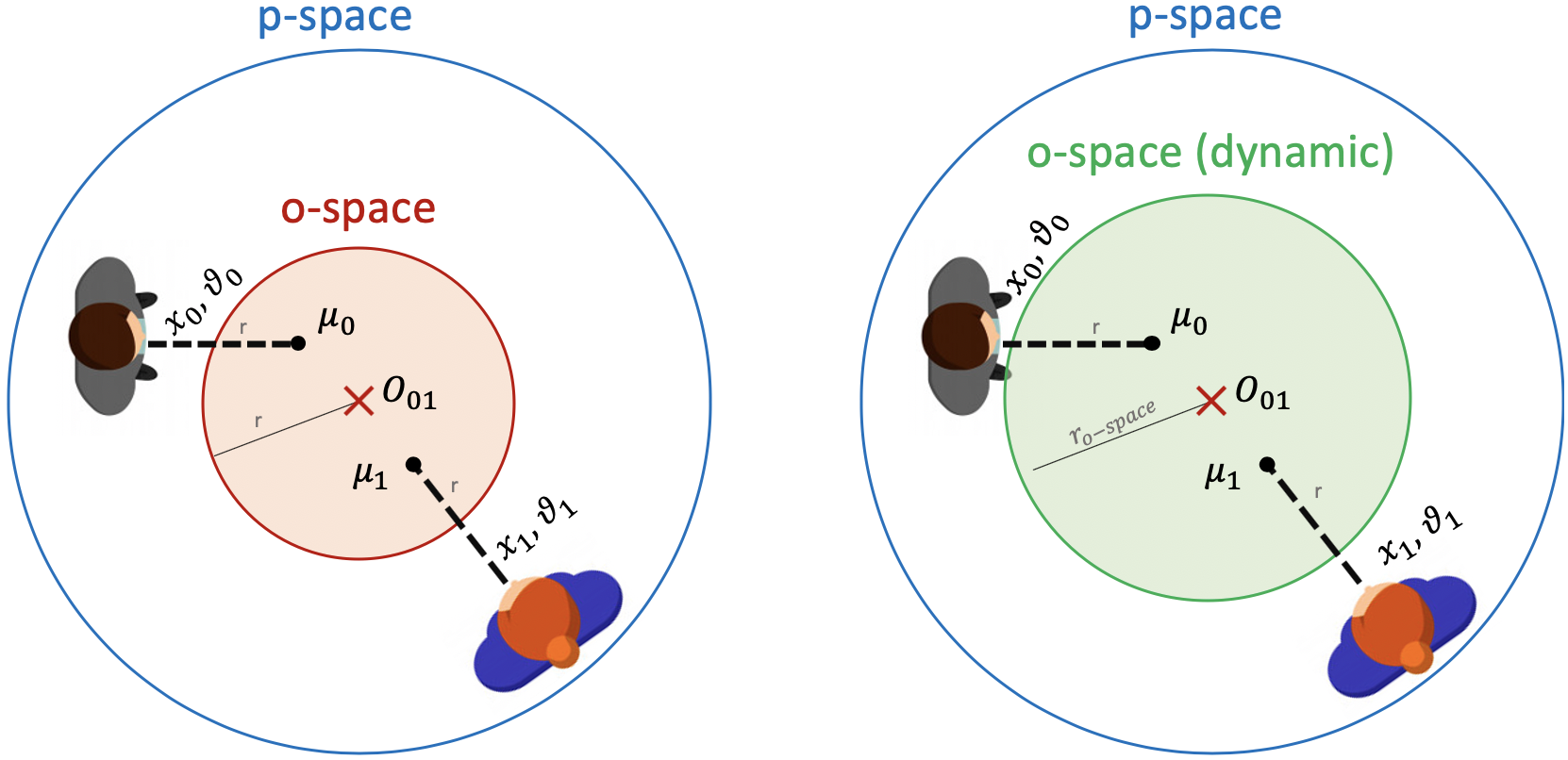}
  \caption{Illustration of the o-space discovery using \cite{cristani2011social} on the left and our approach on the right. Both approaches use the candidate radius $r$ to find the center of the o-space, as infinite number of circles could be drawn from two points. Differently from \cite{cristani2011social}, once a center is found, we dynamically adapt the final radius of the o-space $r_{o-space}$ depending on the effective location of the two people.}
  \label{fig:o-space}
\end{figure}

\subsection{Social Interactions and Distancing}\label{sec:method_si}

We identify social interactions by recognizing the spatial structures that define F-formations (see Section \ref{sec:related_social} for more details). Our approach considers groups of two people in an ``all-vs-all" fashion by studying all the possible pairs of people in an image.

Ideally, two people talking to each other define the same o-space by looking at its center. In practice, 3D localization and orientation of people are noisy and previous methods \cite{cristani2011social, cristani2011towards} have adopted a voting approach. They define a candidate radius \textit{r} of the o-space and each person vote for a center.
The average result defines the center of the o-space. In Cristani \etal \cite{cristani2011social}, the candidate radius $r$ remains the final radius of the o-space and is fixed for every group of people. However, once the o-space center is found, nothing prevents us from considering its radius $r_{o-space}$ dynamically as the minimum distance between the center and one of the two people. An illustration of the differences is show in Figure \ref{fig:o-space}. Therefore, given the location of two people in the x-z plane $\mathbf{x}$ and their body orientation $\theta$, we define the center and the radius of the o-space as:

\begin{equation}
\begin{split}
& \mathbf{O_{01}} = \frac{\boldsymbol{\mu_0} + \boldsymbol{\mu_1}}{2} \\
& r_{o-space} = min( |\mathbf{O_{01}} - \mathbf{x_o}|, | \mathbf{O_{01}} - \mathbf{x_1}|) \;\;\; ,\\
\end{split}
\label{eq:o-space}
\end{equation}
  where $\mathbf{O_{01}}$ and $r_{o-space}$ are the center and radius of the resulting o-space, $\boldsymbol{\mu_0}$ and $\boldsymbol{\mu_1}$ indicate the location of the two candidate centers of the o-space. In general, $\boldsymbol{\mu} = [ x + r*cos(\theta), z + r*sin(\theta) ]$ and is parametrized by the candidate radius $r$, which depends on the type of relation (intimate, personal, business, etc.) \cite{hall1966hidden}. 

Once the o-space is drawn, we verify the following conditions:
\begin{equation}
\begin{split}
(a) \;\; &|\mathbf{x_o} - \mathbf{x_1}| < D_{max} \\
(b) \;\; & |\mathbf{O_{01}} - \mathbf{x_i}| < r_{o-space} \;\; \forall i \ne 0,1 \\ 
(c) \;\; & |\boldsymbol{\mu_0} - \boldsymbol{\mu_1}| < R_{max}
\end{split}
\label{eq:si}
\end{equation}
where $D_{max}$ and $R_{max}$ are the maximum distances between two people, and between the candidate centers of the o-spaces, respectively. Vectors are represented in bold.

The above conditions verify the presence of an F-formation, as:

\begin{enumerate}[label=(\alph*)]
\item examines whether two people stand closer than a maximum distance $D_{max}$, \textit{i.e.}, they lie inside a p-space,
\item examines whether the o-space is empty (no-intrusion condition),
\item examines whether the two people are looking inward the o-space.
\end{enumerate}

We note that condition (c) is empirical as looking inward is a generic requirement. Two people usually look at each other when talking, but the needs for social distancing may be different. Our goal is not to find perfect empirical parameters for F-formations discovery, but rather to show the effectiveness of combining simple rules and estimating 3D localization and orientation.
We consider two people as interacting with each other if the three conditions are verified. This method is automatically extended to larger groups as two people can already cover any possible F-formation (vis-a-vis, L-shape and side-by-side), while three or more people usually form a circle \cite{cristani2011social}. Further, we are not interested in defining the components of each group, but rather whether people are interacting or not.

\vspace{5pt}
\textbf{Social Distancing. } The procedure to monitor social distancing can either follow the same steps, or can be adapted to a different context. Risk of contagion strongly increases if people are involved in a conversation \cite{cvd19airborne, cvd19speaking}. Therefore, recognizing social interactions lets the system only warn those people that incur the highest risk of contagion. In crowded scenes, this is crucial to prevent an extremely high number of false alarms that could undermine any benefit of the technology. Yet social distancing conditions can also be differentiated from the social interaction ones. For example, a third person invading the o-space could mean that the three people involved are not conversing, but still they may be at risk of contagion due to the proximity. How strict these rules should be can only be decided case  by case by the competent authority. Our goal is to help assessing the risk of contagion not only through distance estimation but also by leveraging social cues.

\vspace{5pt}
\textbf{Uncertainty for Social Interactions. }
A deterministic approach can be very sensitive to small errors in 3D localization and orientation, which we know are inevitable due to the perspective projection. 
Therefore, we introduce a probabilistic approach that leverages our estimated uncertainty to increase robustness towards 3D localization noise. We note that Cristani \etal \cite{cristani2011social} also adopted a probabilistic approach injecting uncertainty in a Hough-voting procedure. However, the chosen parameters were driven by sociological and empirical considerations. In our case, uncertainty estimates come directly as an output of the neural network and they are unique for each person.
Recalling that the location of each person is defined as a Laplace distribution parametrized by $d$ and $b$ in Eq. \ref{eq:laplace}, we draw $k$ samples from the distribution. For each pair of samples, we verify the above conditions for social interactions. Combining all the results, we evaluate the final probability for a social interaction to occur.

\section{Experiments}

\begin{figure}[t]
    \centering
    \includegraphics[width=\linewidth]{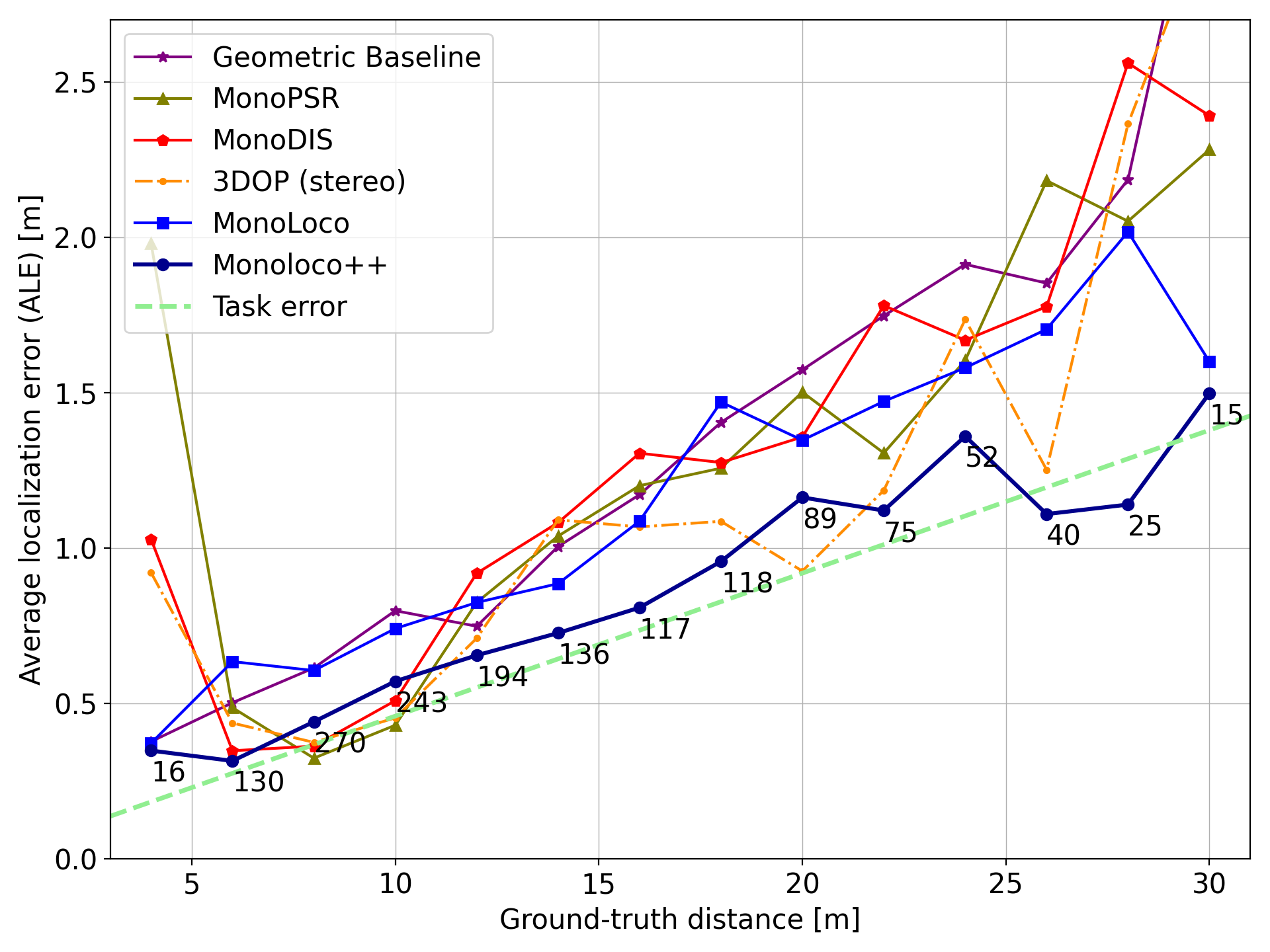}
    \caption{Average localization error (ALE) as a function of distance. We outperform the monocular MonoPSR \cite{ku2019monopsr} and MonoDIS \cite{monodis-tpami}}, while even achieving more stable results than the stereo 3DOP \cite{3dop}. Monocular performances are bounded by our modeled task error in Eq. \ref{eq:task_error}. The task error is only a mathematical construction not used in training and yet it strongly resembles the network error, especially for the more statistically significant clusters (number of predicted instances included).
    \label{fig:results}
    \end{figure}

    \begin{figure}[t]
        \centering
  \includegraphics[width=\linewidth]{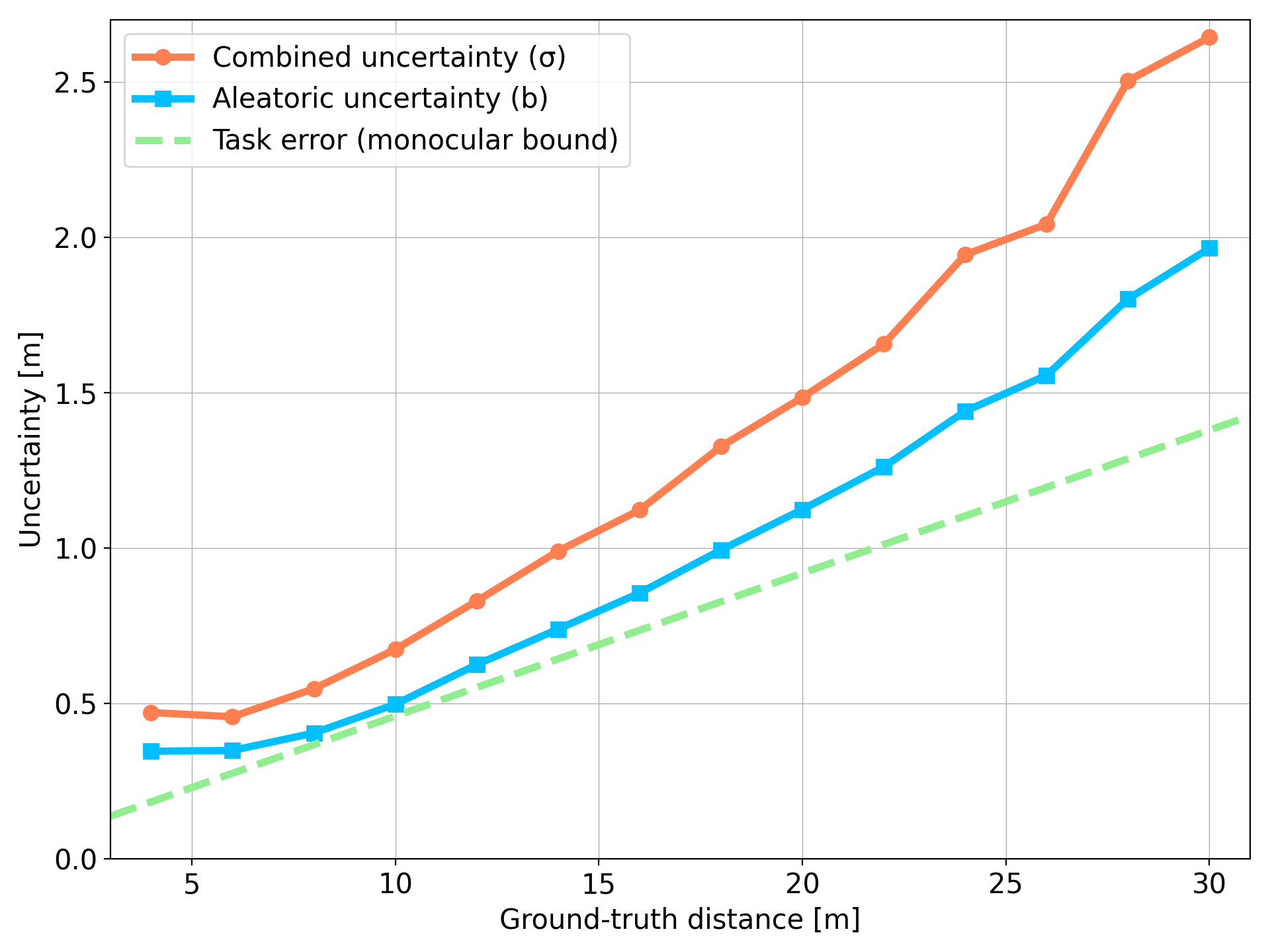}
  \caption{Aleatoric uncertainties predicted by MonoLoco++ (spread $b$),
  and due to human height variation (task error $\hat{e}$) as a function of the ground-truth distance. 
  The term $b - \hat{e}$ is indicative of the aleatoric uncertainty due to noisy observations. The combined uncertainty $\sigma$ accounts for aleatoric and epistemic uncertainty and is obtained applying MC Dropout \cite{Gal2016Dropout} at test time with 50 forward passes.}
      \label{fig:spread}
\end{figure}

To the best of our knowledge, no dataset contains 3D labels as well as social interactions or social distancing information. Hence, we used multiple datasets to evaluate monocular 3D localization, social interactions and social distancing separately. The following sections serve this purpose.

\subsection{Monocular 3D Localization}

\vspace{5pt}
\textbf{Datasets. }
We train and evaluate our monocular model on the KITTI Dataset ~\cite{Geiger2013Kitti}. It contains 7481 training images along with camera calibration files. All the images are captured in the same city from the same camera. To analyze cross-dataset generalization properties, we train another model on the teaser of the recently released \emph{nuScenes} dataset~\cite{nuscenes} and we test it on KITTI. We do not perform cross-dataset training.

\vspace{5pt}
\textbf{Training/Evaluation Procedure. }
To obtain input-output pairs of 2D joints and distances, we apply an off-the-shelf pose detector and use intersection over union of 0.3 to match our detections with the ground-truths, obtaining 1799 training instances for KITTI and 8189 for nuScenes teaser. KITTI images are upscaled by a factor of two to match the minimum dimension of 32 pixels of COCO instances. NuScenes already contains high-definition images, which are not modified. Once the human poses are detected, we apply horizontal flipping to double the instances in the training set.

We follow the KITTI train/val split of Chen \etal \cite{m3d} and we run the training procedure for 200 epochs using the Adam optimizer~\cite{kingma2014adam}, a learning rate of $10^{-3}$ and mini-batches of 512.
The code, available online, is developed using PyTorch~\cite{pytorch}. 
Working with a low-dimensional representation is very appealing as it allows fast experiments with different architectures and hyperparameters. The entire training procedure requires around two minutes on a single GTX1080Ti GPU.

\vspace{5pt}
\textbf{Evaluation Metrics. }
Following \cite{monoloco}, we use two metrics to analyze 3D pedestrian localization. First, we consider a prediction as correct if the error between the predicted distance and the ground-truth is smaller than a threshold. We call this metric Average Localization Accuracy (ALA). We use 0.5 meters, 1 and 2 meters as thresholds.
We also analyze the average localization error (ALE). To make fair comparison we set the threshold of the methods to obtain similar recall. Compared to \cite{monoloco}, we do not evaluate on the common set of detected instances. Their evaluation is not reproducible as the common set depends on the methods chosen for evaluation. In contrast, analyzing ALE and recall allows for simple but fair comparison.
Following KITTI guidelines, we assign to each instance a difficulty regime based on bounding box height, level of occlusion and truncation: \textit{easy}, \textit{moderate} and \textit{hard}. However in practice, each category includes instances from the simpler categories,
and, due to the predominant number of easy instances (1240 \textit{easy} pedestrians, 900 \textit{moderate} and 300 \textit{hard} ones), the metric can be misleading and underestimate the impact of challenging instances. Hence, we evaluate each instance as belonging only to one category and add the category \textit{all} to include all the instances.

\begin{table*}[]
 \centering
  \begin{tabular}{|l|c|c|c c c c|c c c|}
    \hline
    Method & Dataset &  Number of & \multicolumn{4}{c|}{ALE (m) $\downarrow$   \;\; [Recall (\%) $\uparrow$ ]} &  \multicolumn{3}{c|}{ALA (\%) $\uparrow$ } \\
       & Training &  Instances & $Easy$ & $Mod.$  & $Hard$ & $All$
      & $<0.5m$ & $<1m$  & $<2m$ \\
    \hline
    \hline
    Mono3D \cite{m3d}
    & KITTI & 1799 
    & 2.26 [89\%] & 3.00 [65\%] & 3.98 [34\%]& 2.62 [69\%]
    & 13.0 & 22.9 & 38.2 \\
    
    MonoPSR \cite{ku2019monopsr}
    & KITTI & 1799
    & 0.88 [96\%] & 1.86 [68\%] & 1.85 [16\%] & 1.19 [69\%] 
    & 31.1 & 44.2 & 57.4 \\
    
    SMOKE \cite{smoke20}
    & KITTI & 1799 
    & 0.75 [59\%] & 1.30 [30\%] & 1.53 [10\%] & 0.91 [39\%]
    & 18.7 & 27.3 & 34.5 \\
    
    MonoDIS \cite{monodis-tpami}
    & KITTI & 1799 
    &  \textbf{0.66} [85\%] & 1.26 [64\%] & 1.83 [32\%] & 0.93 [66\%] 
    & 33.2 & 47.6 & 57.6 \\
    
    \hline
    3DOP \cite{3dop} (Stereo)
    & KITTI & 1799
    & 0.67 [88\%] & 1.19 [64\%] & 1.93 [37\%] & 0.94 [69\%] 
    & 40.6 & 53.7  & 61.4  \\
    \hline
    
    Geometric \cite{monoloco}
    & KITTI & -
    &  1.05 [89\%] & 0.95 [63\%] & 1.34 [31\%] & 1.04 [68\%]
    & 23.5 & 41.9 & 59.4 \\
    
    MonoLoco \cite{monoloco}
    & KITTI & 1799
    &  0.95 [89\%] & 0.98 [64\%] & 1.11 [31\%] & 0.97 [68\%]
    & 25.3 & 43.4 & 60.5 \\
    
    MonoLoco \cite{monoloco}
    & nuScenes & 8189 
    &  0.91 [92\%] & 1.16 [80\%] & 1.45 [30\%] & 1.08 [74\%]
    & 27.6 & 46.6  & 63.7 \\

    Our MonoLoco++
    & KITTI & 1799
    & 0.69 [90\%] & \textbf{0.71} [66\%] & 1.37 [31\%] & \textbf{0.76} [70\%] 
    & \textbf{37.4} & \textbf{53.2} & 63.6 \\
    
    Our MonoLoco++
    & nuScenes & 1799
    &  0.81 [92\%] & 0.84 [68\%] & 1.14 [29\%] & 0.84 [70\%]
    & 31.8 & 50.2  & 63.9 \\
    
    Our MonoLoco++
    & nuScenes & 8189
    &  0.72 [91\%] & 0.77 [68\%] 
    & \textbf{1.03} [29\%] & \textbf{0.76} [70\%] 
    & 32.5 & 51.9 & \textbf{65.6} \\
    \hline
  \end{tabular}
     \caption{Comparing our proposed method against baseline results on the KITTI dataset ~\cite{Geiger2013Kitti}. 
  We use PifPaf \cite{kreiss2019pifpaf} as off-the-shelf network to extract 2D poses. For the ALE metric, we show the recall between brackets to insure fair comparison. We show results by training with three different data splits: KITTI dataset \cite{Geiger2013Kitti}, nuScenes teaser \cite{nuscenes} or a subset of nuScenes to match the number of instances of the KITTI dataset. All cases share the same evaluation protocol. The models trained on nuScenes show cross-dataset generalization properties by obtaining comparable results in the ALE metric.}
  \label{tab:res_kitti}
\end{table*}
\vspace{5pt}
\textbf{Geometric Approach. } 
3D pedestrian localization is an ill-posed task due to human height variations.  
On the other side, estimating the distance of an object of known dimensions from its projections 
into the image plane is a well-known deterministic problem. As a baseline, we consider humans as fixed objects 
with the same height and we investigate the localization accuracy under this assumption.

For every pedestrian, we apply a pose detector to calculate distances in pixels between different body parts in the image domain. 
Combining this information with the location of the person in the world domain, we analyze the distribution of the real dimensions (in meters) of all the instances in the training set for three segments: 
head to shoulder, shoulder to hip and hip to ankle.
For our calculation we assume a pinhole model of the camera and that all instances stand upright.
Using the camera intrinsic matrix K and knowing the ground-truth location of each instance 
$ \textbf{D} = \left[x_c, y_c, z_c \right]^T $ we can back-project each keypoint from the image plane to its 3D 
location and measure the height of each segment using Eq. \ref{eq:k}. 
We calculate the mean and the standard deviation in meters of each of the segments for all the instances in the training set. 
The standard deviation is used to choose the most stable segment for our calculations. 
For instance, the position of the head with respect to shoulders may vary a lot for each instance. 
We also average between left and right keypoint values to take into account noise in the 2D joint predictions. 
The result is a single height $\Delta y_{1-2}$ that represents the average length of two body parts. 
In practice, our geometric baseline uses the \textit{shoulder-hip} segment and predicts an average height of $50.5cm$. Combining the study on human heights \cite{visscher2008sizing} described in Section 3  with the anthropometry study of Drillis \etal \cite{drillis1969body}, we can compare our estimated $\Delta y_{1-2}$ with the human average \textit{shoulder-hip} height: $0.288 * 171.5cm = 49.3cm$.

The next step is to calculate the location of each instance knowing the value in pixels of the chosen keypoints  $v_1$ and $v_2$ and assuming $\Delta y_{1-2}$ to be their relative distance in meters. 
This configuration requires to solve an over-constrained linear system with two specular solutions, of which only one is inside the camera field of view.

\vspace{5pt}
\textbf{Other Baselines. }
We compare our monocular method on KITTI against five monocular approaches and a stereo one:
\begin{itemize}
    \item \textit{MonoLoco}. We compare our approach with MonoLoco \cite{monoloco}. Our MonoLoco++ uses a multi-task approach to learn orientation, has a different architecture and uses spherical coordinates for distance estimation. Both methods share the same off-the-shelf pose detector \cite{kreiss2019pifpaf, kreiss2021openpifpaf}.
   \item \textit{Mono3D} \cite{m3d} is a monocular 3D object detector for cars, cyclists and pedestrians. 3D localization of pedestrians is not evaluated but detection results are publicly available
   \item \textit{MonoPSR} \cite{ku2019monopsr} is a monocular 3D object detector that leverages point clouds at training time to learn shapes of objects. In contrast, our method does not use any privileged signal at training time.
    \item \textit{MonoDIS} \cite{monodis-tpami} is a very recent multi-class 3D object detector that provides evaluations for the pedestrian category on the KITTI dataset.
    \item \textit{SMOKE} \cite{smoke20} is a single-stage monocular 3D object detection method which is based on projecting 3D points onto the image plane. The authors have shared their quantitative evaluation.
   \item \textit{3DOP} \cite{3dop} is a stereo approach for pedestrians, cars and cyclists and their 3D detections are publicly available.
\end{itemize}
Finally, in Figure \ref{fig:results} we also compare the results against the task error of Eq.~\ref{eq:task_error}, which defines the target error for monocular approaches due to the ambiguity of the task.

\begin{table}
  \centering
  \begin{tabular}{|l| c c c|}
    \hline
    & $|x - d|/ \sigma$ &$ |\sigma - e|$ [m] & Recall [\%]\\
    \hline
    \hline
    $p_{drop}=0.05 $  & 0.60 & 0.90 & 82.8  \\
    $p_{drop}=0.2 $  & 0.58 & 0.96 & 84.3 \\
    $p_{drop}=0.4 $  & 0.50 & 1.26 & 88.3 \\
    \hline
  \end{tabular}
  \caption{Precision and recall of uncertainty for the KITTI validation set with 50 stochastic forward passes. $|x- d|$ is the localization error, $\sigma$ the predicted confidence interval, $\hat{e}$ the task error modeled in Eq. \ref{eq:task_error} and Recall is represented by the \% of ground-truth instances inside the predicted confidence interval.}
  \label{tab:uncertainty}
\end{table}

\begin{table}
  \centering
  \begin{tabular}{|l|c c c c c|}
    \hline
    
    Mask R-CNN  & & & ALE [m] & & \\
    \cite{He2017MaskR} & ${_0^{10}}$ & ${_{10}^{20}}$ & ${_{20}^{30}}$ & ${_{30}^+}$ & ${All}$ \\
    \hline\hline
    Geometric
    & 0.79 & 1.52 & 3.17 & 9.08 & 3.73 \\
    \hline
    $L_1$ loss
    & 0.85 & \textbf{1.17} & 2.24 & 4.11 & 2.14 \\
    Gaussian loss
    & 0.90 & 1.28 & 2.34 & 4.32 & 2.26  \\
    Laplace Loss
    & \textbf{0.74} & \textbf{1.17} & 2.25 & 4.12 & 2.12  \\
    \hline
    \hline
    PifPaf \cite{kreiss2019pifpaf} & & & ALE [m] & & \\
    & ${_0^{10}}$ & ${_{10}^{20}}$ & ${_{20}^{30}}$ & ${_{30}^+}$ & ${All}$ \\
    \hline\hline
    Geometric
    & 0.83 & 1.40 & 2.15 & 3.59 & 2.05 \\
    \hline
    $L_1$ loss
    & 0.83 & 1.24 & \textbf{2.09} & 3.32 & 1.92 \\
    Gaussian loss
    & 0.89 & 1.22 & 2.14 & 3.50 & 1.97 \\
    \textbf{Laplace loss}
    & 0.75 & 1.19 & 2.24 & \textbf{3.25} & \textbf{1.90} \\
    \hline
  \end{tabular}
  \caption{Impact of different loss functions with Mask R-CNN \cite{He2017MaskR} and PifPaf \cite{kreiss2019pifpaf} pose detectors on nuScenes teaser validation set \cite{nuscenes}. We also show results using the Average Localization Error (ALE) metric as a function of the ground-truth distance using clusters of 10 meters.}
  \label{tab:ablation}
\end{table}

\begin{table}
  \centering
\resizebox{\columnwidth}{!}{
  \begin{tabular}{|l| c c c|}
    \hline
    Method $ \setminus $ Time [ms] & $t^{pose}$ & $t^{model}$& $t^{total}$\\
    \hline
    \hline
    Mono3D \cite{m3d} & - & 1800 & 1800   \\
    3DOP \cite{3dop} & - & $~$2000 & $~$2000 \\
    MonoPSR \cite{ku2019monopsr} & - & $~$200 & $~$200 \\
    Our MonoLoco++ (1 sample) & 89 / 162 & 10  & \textbf{99 / 172} \\
    Our MonoLoco++ (50 samples) & 89 / 162 & 51 & 140 / 213\\
    \hline
  \end{tabular}
  }
  \caption{ Single-image inference time on a single GTX 1080Ti for the KITTI dataset \cite{Geiger2013Kitti} with PifPaf \cite{kreiss2019pifpaf} as pose detector. Most computation comes from the pose detector (ResNet 50 / ResNet 152 backbones). For this study, we use all the images at their original scale that contain at least a pedestrian. For Mono3D, 3DOP and MonoPSR we report published statistics on a Titan X GPU.
  In the last line, we calculate epistemic uncertainty through 50 sequential forward passes. In future work, this computation can be parallelized.
 }
  \label{tab:runtime}
\end{table}

\subsection{Monocular Results}

\vspace{5pt}
\textbf{Localization Accuracy. } 
Table \ref{tab:res_kitti} summarizes our quantitative results on KITTI. 
We strongly outperform all the other monocular approaches on all metrics and obtain comparable results with the stereo approach 3DOP  \cite{3dop}, which
has been trained and evaluated on KITTI 
and makes use of stereo images during training and test time. In addition, we show cross-dataset generalization properties by training our network on a subset of the nuScenes dataset containing only 1799 instances and evaluating it on the KITTI dataset. Its generalization properties can be attributed to the low-dimensional input space of 2D keypoints \cite{virtualsamples19}.

\begin{figure*}[t]
  \centering
    \includegraphics[width=0.95\linewidth]{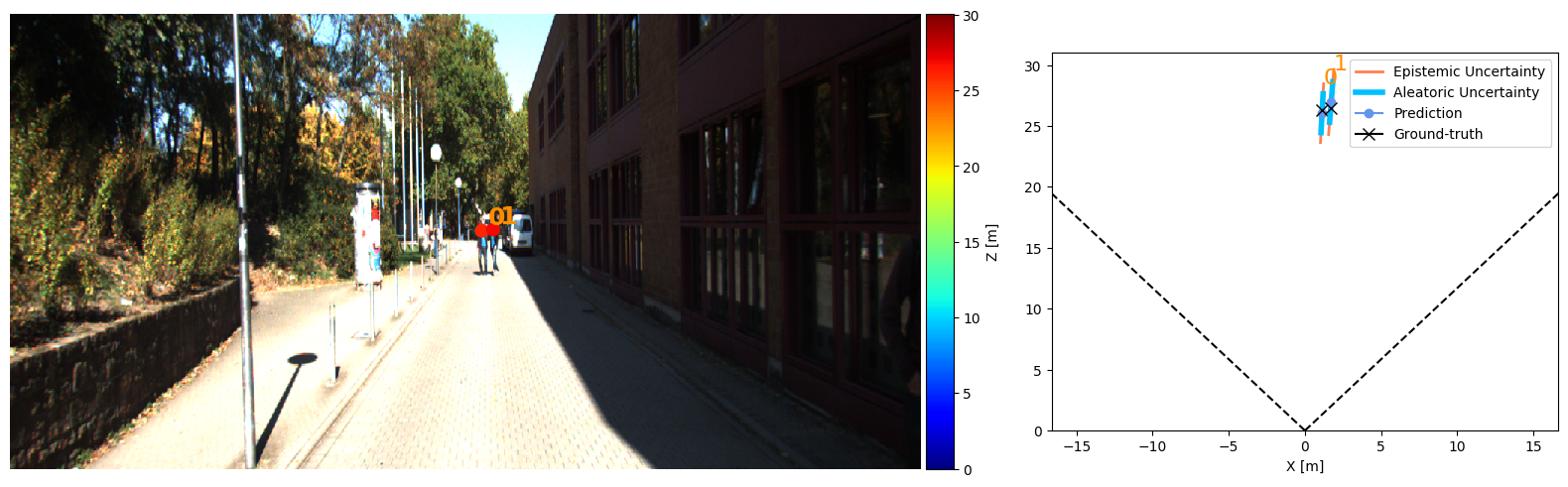}
    \includegraphics[width=0.95\linewidth]{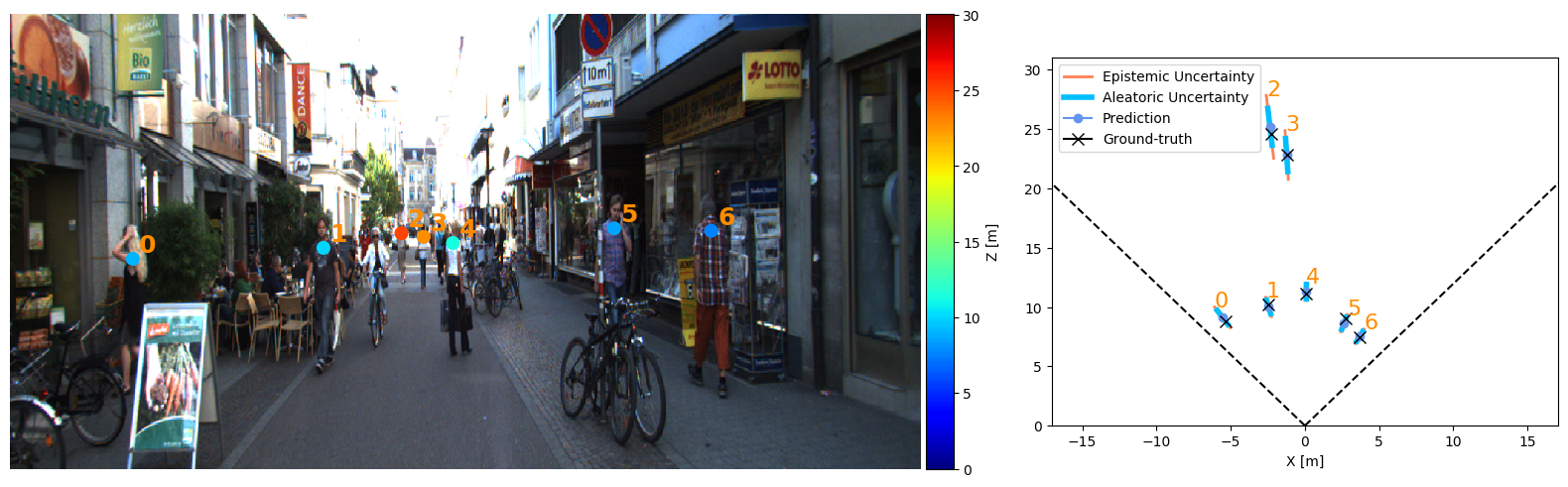}
    \caption{Qualitative results from the KITTI ~\cite{Geiger2013Kitti} dataset containing true and inferred distance information as well as confidence intervals. The direction of the line is radial as we use spherical coordinates. Only pedestrians that match a ground-truth are shown for clarity.}
  \label{fig:qual_ki}
\end{figure*}

\begin{figure*}[t]
  \centering
    \includegraphics[width=0.95\linewidth]{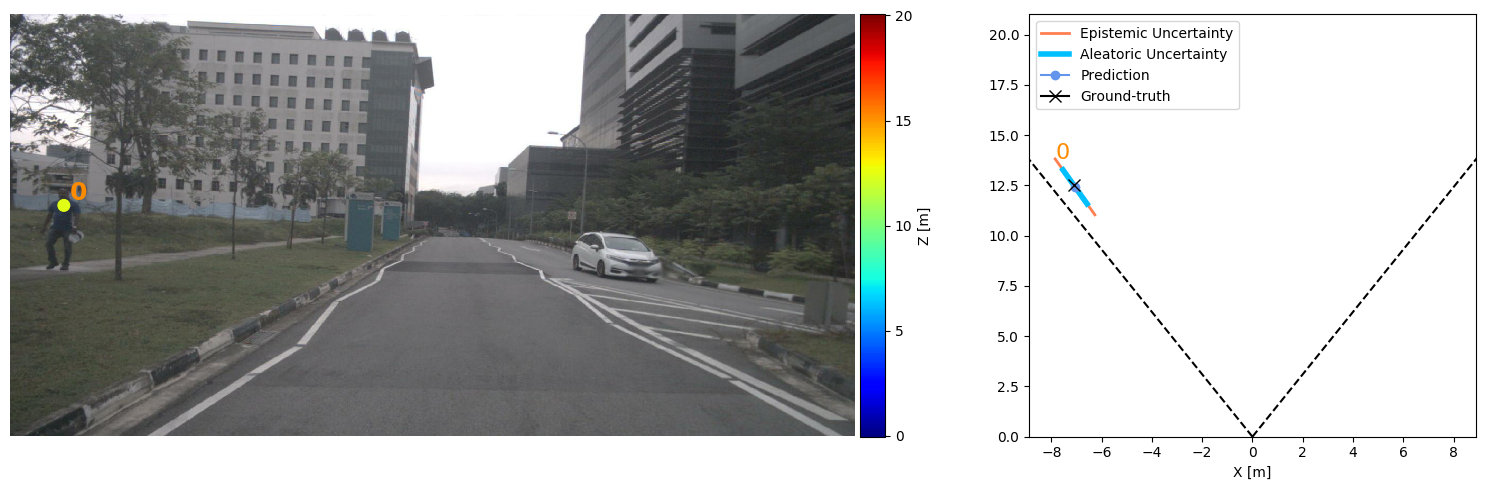}
    \includegraphics[width=0.95\linewidth]{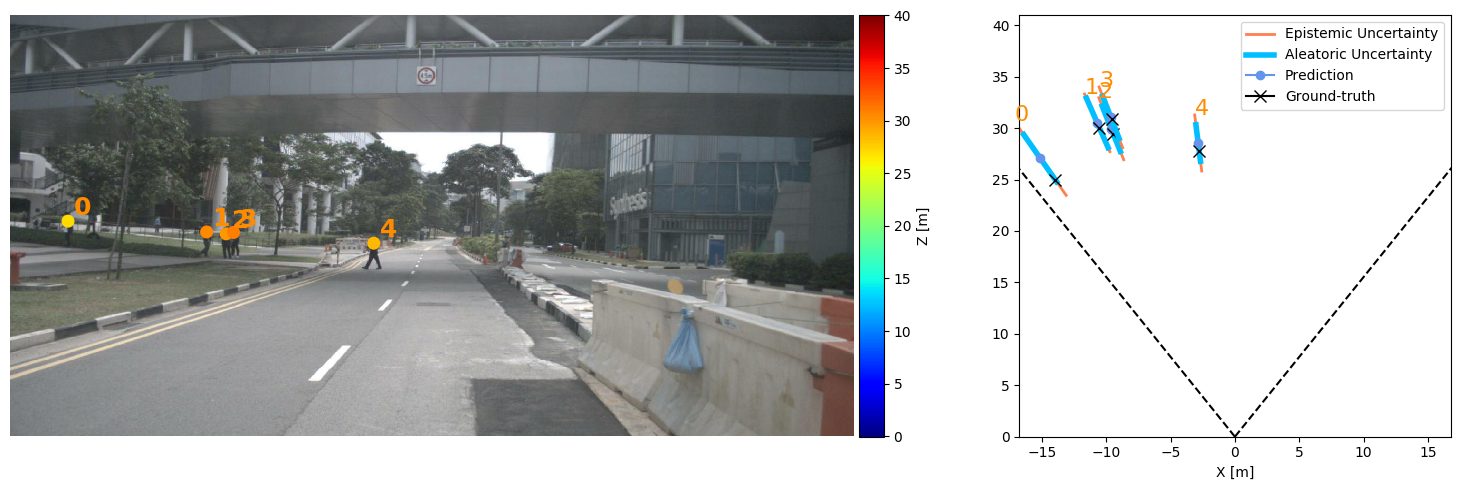}
    \includegraphics[width=0.95\linewidth]{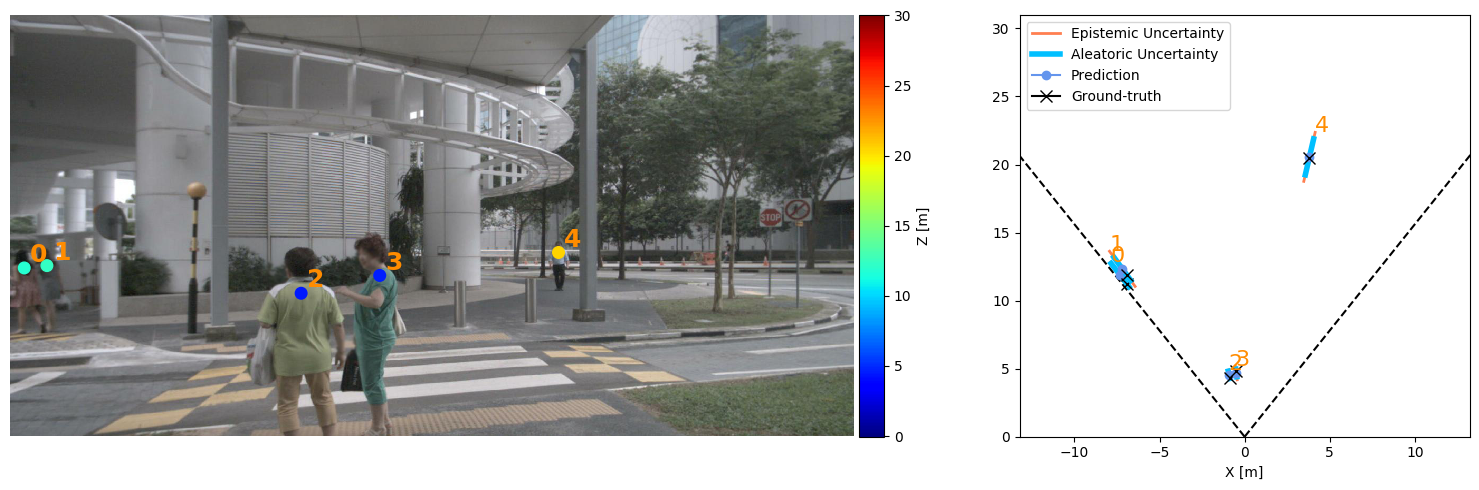}
    \caption{Qualitative results from the nuScenes dataset \cite{nuscenes} containing true and inferred distance information as well as confidence intervals.}
  \label{fig:qual_nu}
\end{figure*}

In Figure \ref{fig:results}, we make an in-depth comparison analyzing the average localization error as a function of the ground-truth distance. We also compare the performances against the \textit{task error} due to human height variations modeled in equation \ref{eq:task_error}.
Our method results in stable performances that almost replicate the target threshold. More generally, it is notable that the error of each method shows a quasi-linear behaviour. At a short range, the majority of methods show large errors, as the instances are not fully visible in the image. Since our method reasons with keypoints, its performances are more stable. At  the 25-30m range MonoLoco error is slightly lower than the task error. This is mainly caused by the statistical fluctuations due to the small sample sizes at those distances.
Figure \ref{fig:qual_ki} and \ref{fig:qual_nu} show qualitative results on challenging images from the KITTI and nuScenes datasets, respectively.

\vspace{5pt}
\textbf{Aleatoric uncertainty. }
We compare in Figure \ref{fig:spread} the aleatoric uncertainty predicted by our network through spread \textit{b} with the \textit{task error} due to human height variation defined in Eq. \ref{eq:task_error}.
While $\hat{e}$ is a linear function of the distance from the camera, the predicted aleatoric uncertainty (through the spread $b$) is a property of each set of inputs. In fact, $b$ includes  not only the uncertainty due to the ambiguity of the task but also the uncertainty due to noisy observations \cite{Kendall2017WhatUD}, \ie, the 2D joints inferred by the pose detector.
Hence, we can approximately define the predictive aleatoric uncertainty due to noisy joints as $b - \hat{e}$ 
and we observe that the further a person is from the camera, the higher is the term $b - \hat{e}$.
The spread $b$ is the result of a probabilistic interpretation of the model and the resulting confidence intervals are calibrated. On the KITTI validation set, they include 68\% of the instances. 

\vspace{5pt}
\textbf{Combined Uncertainty. }
The combined aleatoric and epistemic uncertainties are captured by sampling from multiple Laplace distributions using MC dropout. During each of the forward passes, we draw and accumulate samples from the estimated Laplace distribution. Then, we calculate the combined uncertainty as the sample variance of predicted distances in Eq. \ref{eq:variance}. The magnitude of the uncertainty depends on the chosen dropout probability $p_{\textrm{drop}}$ in Eq. \ref{mc_drop}.
In Table \ref{tab:uncertainty}, we analyze the precision/recall trade-off for different dropout probabilities and choose $p_{\textrm{drop}} = 0.2$. We perform 50 computationally expensive forward passes and, for each of them, 100 computationally cheap samples from a Laplace distribution using Eq.~\ref{eq:variance}. As a result, $84\%$ of pedestrians lie inside the predicted confidence intervals for the validation set of KITTI.

One of our goals is robust 3D estimates for pedestrians, and being able to predict a confidence interval instead of a single regression number is a first step towards this direction. 
To illustrate the benefits of predicting intervals over point estimates, we construct a controlled risk analysis. 
To simulate an autonomous driving scenario,
we define as \textit{high-risk cases} all those instances where the ground-truth distance is smaller than the predicted one, 
hence a collision is more likely to happen. 
We estimate that among the 1932 detected pedestrians in KITTI which match a ground-truth, 48\% of them are considered as \textit{high-risk cases}, 
but for 89\% of them the ground-truth lies inside the predicted interval. 

\begin{figure*}[t]
  \centering
    \includegraphics[width=0.95\linewidth]{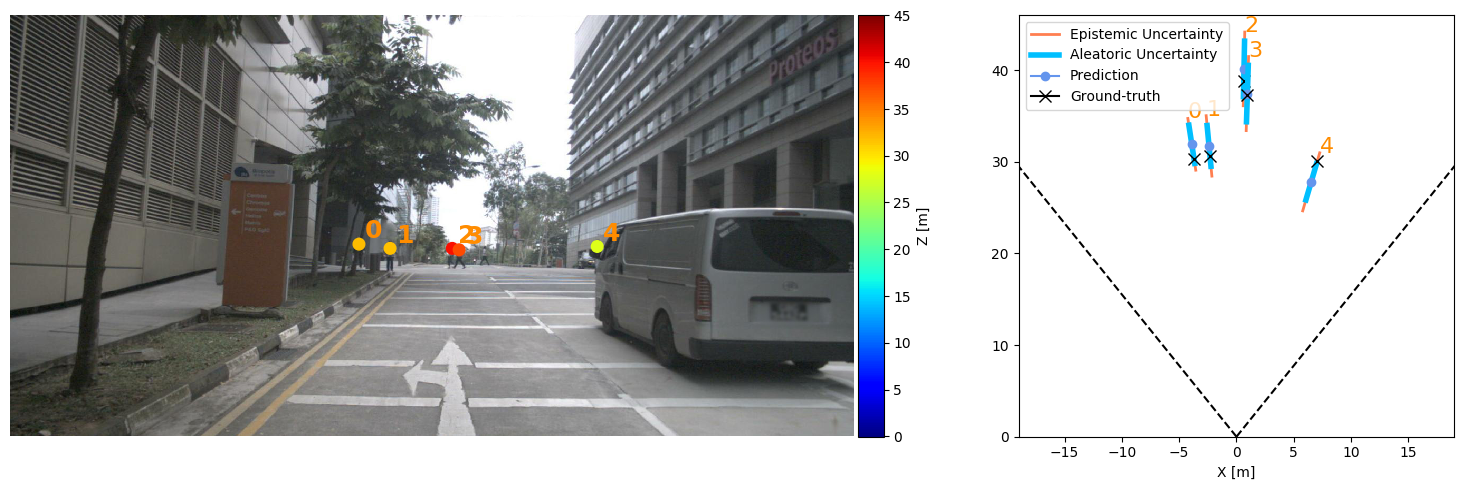}
    \includegraphics[width=0.95\linewidth]{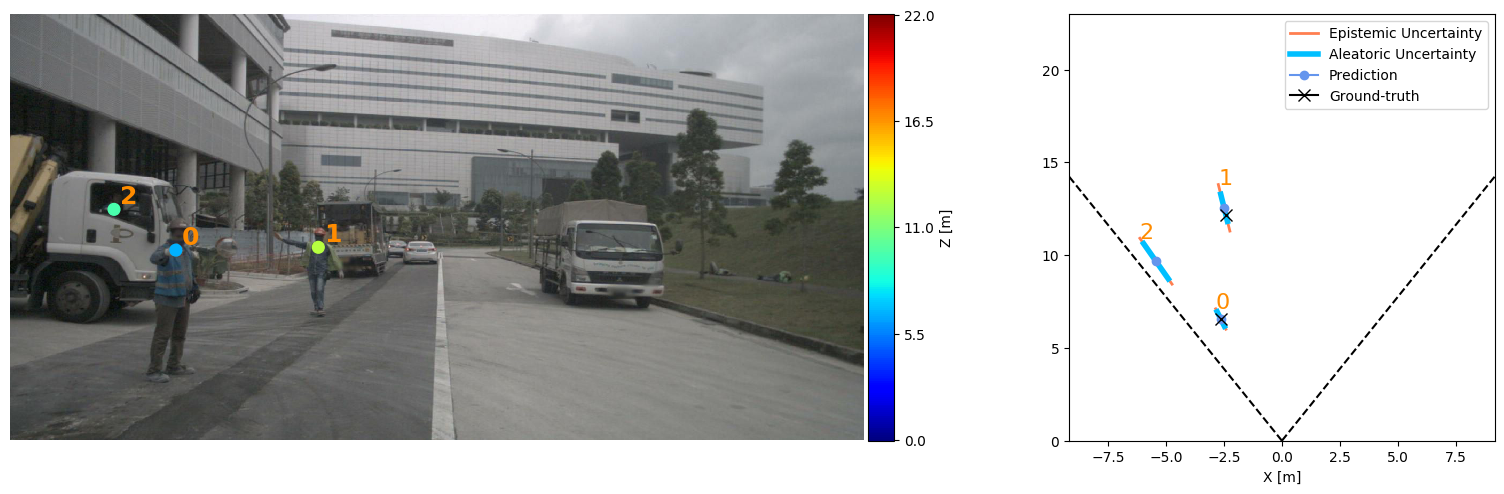}
    \caption{These examples show 1) why relying on homography or assuming a flat plane can be dangerous, and 2) the importance of uncertainty estimation. In the top image, the road is uphill and the assumption of a constant flat plane would not stand. MonoLoco++ accurately detects people up to 40 meters away. Instance 4 is partially occluded by a van and this is reflected in a higher uncertainty. In the bottom image, we also detect a person inside a truck. No ground-truth is available for the driver but empirically the prediction looks accurate. Furthermore, the estimated uncertainty increases, which is a useful indicator to warn about critical samples. } 
  \label{fig:qual_challenge}
\end{figure*}

\vspace{5pt}
\textbf{Challenging Cases. }
We qualitatively analyze the role of the predicted uncertainty in case of an outlier in Figure \ref{fig:qual_challenge}. In the top image, a person is partially occluded and this is reflected in a larger confidence interval. Similarly in the bottom figure, we estimate the 3D localization of a driver inside a truck. The network responds to the unusual position of the 2D joints with a very large confidence interval. In this case the prediction is also reasonably accurate, but in general an unusual uncertainty can be interpreted as a useful indicator to warn about critical samples. 

We also show the advantage of estimating distances without relying on homography estimation or assuming a fixed ground plane, such as \cite{m3d, 3dop}. The road in Figure \ref{fig:qual_challenge} (top) is uphill as frequently happens in the real world (\eg, San Francisco). MonoLoco++ does not rely on ground plane estimation, making it robust to such cases.

\vspace{5pt}
\textbf{Ablation Studies. }
In Table \ref{tab:ablation}, we analyze the effects of choosing a top-down or a bottom-up pose detector with different loss functions and with our deterministic geometric baseline. We compare our Laplace-based $L1$ loss of Eq. \ref{eq:laplace} with a relative $L_1$ loss

\begin{equation}
  L_{\textrm{1}}(x|d) = |1-d/x| \;\;\; ,
\label{eq:l1} 
\end{equation}
and a Gaussian loss
\begin{equation}
  L_{\textrm{Gaussian}}(x|d,\sigma) = \frac{(1-d/x)^2}{2\sigma^2} + \frac{1}{2}\log(\sigma^2) \;\;\; .
\label{eq:gaussian} 
\end{equation}

The Gaussian Loss is based on the negative log-likelihood of a Gaussian distribution  and corresponds to an $L_2$ loss attenuated by a predicted $\sigma$ in the location. Intuitively, $L_2$ type losses are more sensitive to outliers due to their quadratic component. All the losses make use of relative distances for consistency with Eq. \ref{eq:laplace}. From Table  \ref{tab:ablation}, we observe that
$L_1$-type losses perform slightly better than the Gaussian loss, but the main improvement is given by choosing PifPaf as pose detector.

\vspace{5pt}
\textbf{Run Time. }
A run time comparison is shown in Table \ref{tab:runtime}. Our method is faster or comparable to all the other methods, achieving real-time performance.

\subsection{Social Interactions}
To evaluate social interactions we focus on the activity of \textit{talking}, which is considered as the most common form of social interaction \cite{cristani2011social}. From single images, we evaluate how well we recognize whether people are talking or just passing by, walking away etc.

\vspace{5pt}
\textbf{Datasets. } We evaluate social interactions on the  Collective Activity Dataset \cite{choi2009they}, which contains 44 video sequences of 5 different collective activities: \textit{crossing}, \textit{walking}, \textit{waiting}, \textit{talking}, and \textit{queuing} and focus on the \textit{talking} activity. The \textit{talkin}g activity is recorded for both indoor and outdoor scenes, allowing us to test our 3D localization performance on different scenarios. Compared to other deep learning methods \cite{caetano2019skeleton,bagautdinov2017social,gavrilyuk2020actor}, we analyze each frame independently with no temporal information, and we do not perform any training for this task, using all the dataset for testing.

\begin{table}
  \centering
  \begin{tabular}{|l|c|c|}
    \hline
    Method  & Accuracy (\%) $\uparrow$  & Recall (\%)$\uparrow$ \\
    \hline
    \hline
    W/o Orientation & 67.0 & 97.2 \\
    Deterministic & 83.7 & 97.2\\
    Task Error Uncertainty & 91.3  & 97.2 \\
    MonoLoco++ Uncertainty & \textbf{91.5}  & 97.2 \\
    \hline
  \end{tabular}
  \caption{Accuracy in recognizing the \textit{talking} activity on the Collective Activity dataset \cite{choi2009they}. In all cases the distance has been estimated by our MonoLoco++. ``W/o Orientation", does not uses the estimated orientation, while ``Deterministic" leverages orientation but not the uncertainty. ``Task Error Uncertainty" refers to the distance-based uncertainty due to ambiguity in the task (Eq. \ref{eq:task_error}), ``MonoLoco++ Uncertainty" refers to the instance-based uncertainty estimated by our MonoLoco++.}
  \label{tab:probabilistic}
\end{table}

\vspace{5pt}
\textbf{Evaluation. }
For each person in the image, we estimate his/her 3D localization confidence interval and orientation. For every pair of people we apply Eq. \ref{eq:o-space} and Eq. \ref{eq:si} to discover the F-formation and assess its suitability. We use the following parameters in meters: $D_{max} = 2$ as maximum distance, and $r_1=0.3$, $r_2=0.5$ and $r_3=1$ as radii for o-space candidates. These choices reflect the average distances of \textit{intimate relations}, \textit{casual/personal relations} and \textit{social/consultive} relations, respectively \cite{hall1966hidden}. 

How much people should look inward the o-space (to assume they are talking) is also an empirical evaluation. We set the maximum distance between two candidate centers $R_{max}=r_{o-space}$ for simplicity. We treat the problem as a binary classification task and evaluate the detection recall and the accuracy in estimating whether the detected people are talking to each other. To disentangle the role of the 2D detection task, we report accuracy on the instances that match a ground-truth. To avoid class imbalance, we only analyze sequences that contain at least a person talking in one of their frames. Consequently, we evaluate a total of 4328 instances, of which 52.8\% are \textit{talking}.

\vspace{5pt}
\textbf{Voting Procedure. }
To account for noise in 3D localization, we sample our results from the estimated Laplace distribution parameterized by distance $d$ and spread $b$ (Eq. \ref{eq:laplace}). Each sample votes for a candidate center $\boldsymbol{\mu}$ and we accumulate the voting. If an agreement is reached within at least 25\% of the samples, we consider the target pair of people as involved in a social interaction and/or at risk of contagion. 
MonoLoco++ estimates a unique spread $b$ for each pedestrian, which accounts for occlusions or unusual locations, as seen in Figure \ref{fig:qual_challenge}. Further, we compare this technique to (i) a baseline approach that leverages 3D localization but not orientation, (ii) a deterministic approach that does not include uncertainty, and (iii) a probabilistic approach where the uncertainty is provided by the task error defined in Eq. \ref{eq:task_error}.

\begin{figure*}[t]
  \centering
    \includegraphics[width=0.95\linewidth]{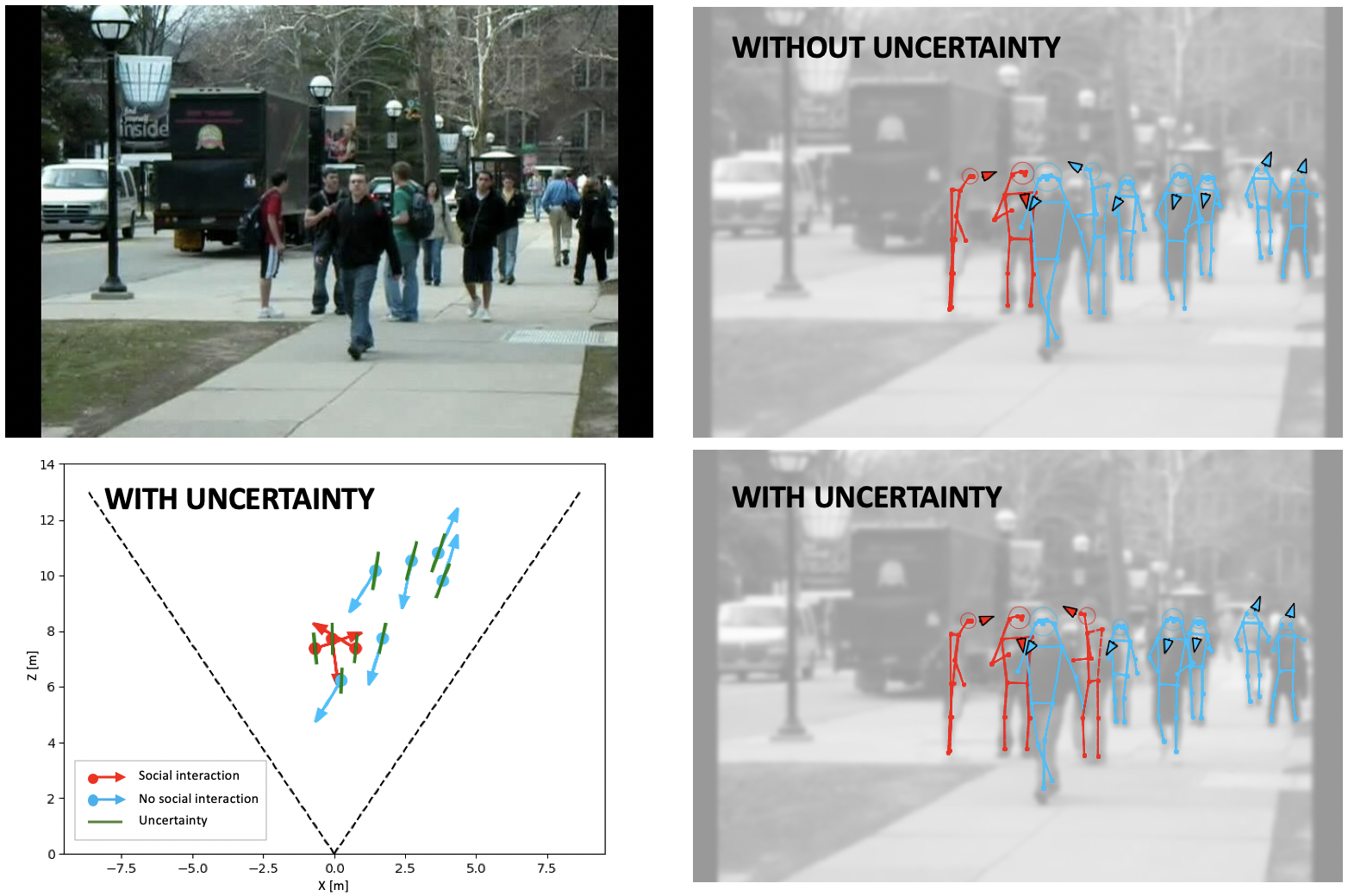}
    \caption{Estimating whether people are talking to each other (social interaction). The use of uncertainty makes the method more robust to 3D localization errors and improves the accuracy. The bird eye view shows the estimated 3D locations and orientations of all the people. The color of the arrows indicates whether people are involved in talking.} 
  \label{fig:si1}
\end{figure*}

\begin{figure*}[t]
  \centering
    \includegraphics[width=0.95\linewidth]{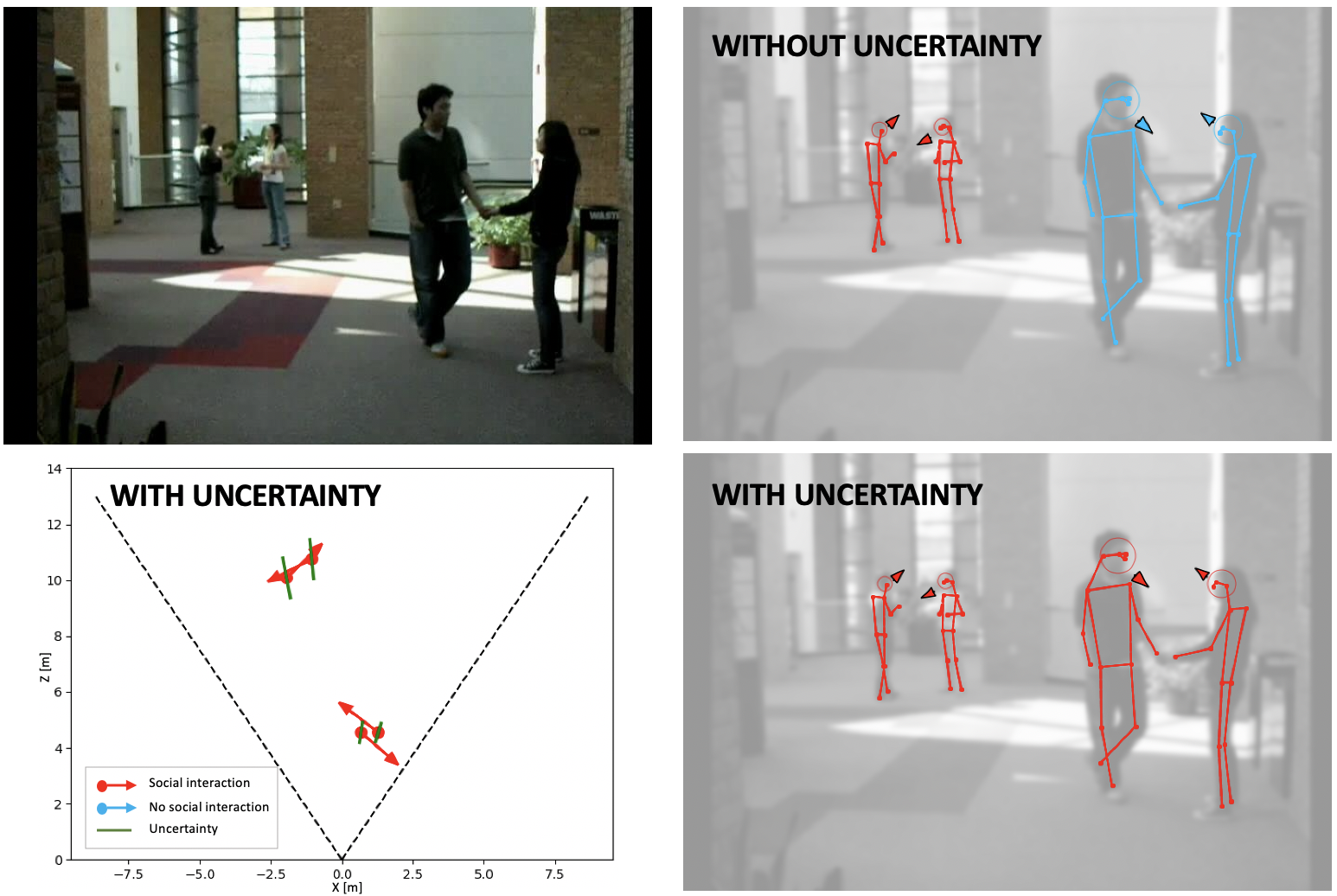}
    \caption{Estimating whether people are talking to each other(social interaction). Even small errors in 3D localization can lead to wrong predictions. As shown in the bird eye view, the estimated locations of the two people is only slightly off due to the height variation of the subjects. Uncertainty estimation compensates the error due to the ambiguity of the task.} 
  \label{fig:si2}
\end{figure*}

\vspace{5pt}
\textbf{Results. } 
Table \ref{tab:probabilistic} shows the results for the\textit{ talking} activity in the Collective Activity Dataset \cite{choi2009they}. Our MonoLoco++ detects whether people are talking from a single RGB image with 91.5\% accuracy without being trained on this dataset, but only using the estimated 3D localization and orientation. The uncertainty estimation plays a crucial role in dealing with noisy 3D localizations as shown in the ablation study of Table~\ref{tab:probabilistic}. All approaches use the same values for 3D localization and orientation, but they differ in their uncertainty component. The biggest improvement is given from deterministic approaches (Row 1, Row 2) to a probabilistic one. Row 3 refers to the task error uncertainty of Eq. \ref{eq:task_error}, which grows linearly with distance. Row 4 refers to the estimated confidence interval from MonoLoco++, which are unique for each person. The role of uncertainty is also shown in Figures~\ref{fig:si1} and \ref{fig:si2},  where 3D localization errors are compensated by the voting procedure.

\begin{table}[]
 \centering
  \begin{tabular}{|l|c c c c|c c c|}
    \hline
    Method & \multicolumn{4}{c|}{Accuracy (\%) $\uparrow$   \;\; [Recall (\%) $\uparrow$ ]}\\
    & $Easy$ & $Mod.$  & $Hard$ & $All$\\
    \hline
    \hline
    W/o Orientation
    & 84.0 [95] & 80.9 [75] & 82.5 [33] & 83.3 [75]\\
    Deterministic
    & 80.5 [95] & 77.9 [75] & 79.0 [33] & 79.8 [75]\\
    Task Error U.
    & 84.2 [95] & 81.4 [75] & 85.3 [33] & 83.6 [75]\\
    MonoLoco++ U.
    &  \textbf{84.7} [95] & \textbf{81.6} [77] 
    & \textbf{85.3} [33]& \textbf{84.0} [75] \\
    \hline
  \end{tabular}
     \caption{Accuracy in monitoring social distancing on KITTI dataset \cite{Geiger2013Kitti}. In all cases the distance has been estimated by our MonoLoco++. ``W/o Orientation", does not uses orientation to account for social distancing, while ``Deterministic" leverages orientation but not the uncertainty. ``Task Error U." refers to the distance-based uncertainty due to ambiguity in the task (Eq. \ref{eq:task_error}), ``MonoLoco++ U." refers to the instance-based uncertainty estimated by our MonoLoco++.}
  \label{tab:social_distance}
\end{table}

\subsection{Social Distancing}
Regarding social distancing, there are no fixed rules for evaluation. As previously discussed, the risk of contagion is higher when people are talking to each other \cite{asadi2019aerosol}, yet it may be necessary to maintain social distancing also when people are simply too close. Our goal is not to provide effective rules, but a framework to assess whether a given set of rules is respected. 

\vspace{5pt}
\textbf{Datasets. }
In the absence of a dataset for social distancing, we created one by augmenting 3D labels of the KITTI dataset~\cite{Geiger2013Kitti}. We apply Eq. \ref{eq:si} using the ground-truth localization and orientation to define whether people are violating social distancing. Once every person is assigned a binary attribute, we evaluate our accuracy on this classification task using our estimated 3D localization and orientation and applying the same set of rules.

\vspace{5pt}
\textbf{Evaluation. }
We evaluate on the augmented KITTI dataset where every person has been assigned a binary attribute for social distancing. Coherently with the monocular 3D localization task, we evaluate on the val split of Chen \etal \cite{m3d} even if no training is performed for this task.  We use the same parameters as for the social interaction task, only relaxing the constraint on how people should look inward the o-space, and we set $R_{max} = 2* r_{o-space}$. This corresponds to verifying whether both candidate centers $\boldsymbol{\mu_0}, \boldsymbol{\mu_1}$ are inside the o-space, as shown in Figure \ref{fig:o-space}. The larger $R_{max}$ in Eq. \ref{eq:si}c, the more conservative the social distancing requirement. If Eq. \ref{eq:si}c is removed completely, social distancing would only depend on the distance between people. 

\begin{figure*}[t]
  \centering
    \includegraphics[width=0.95\linewidth]{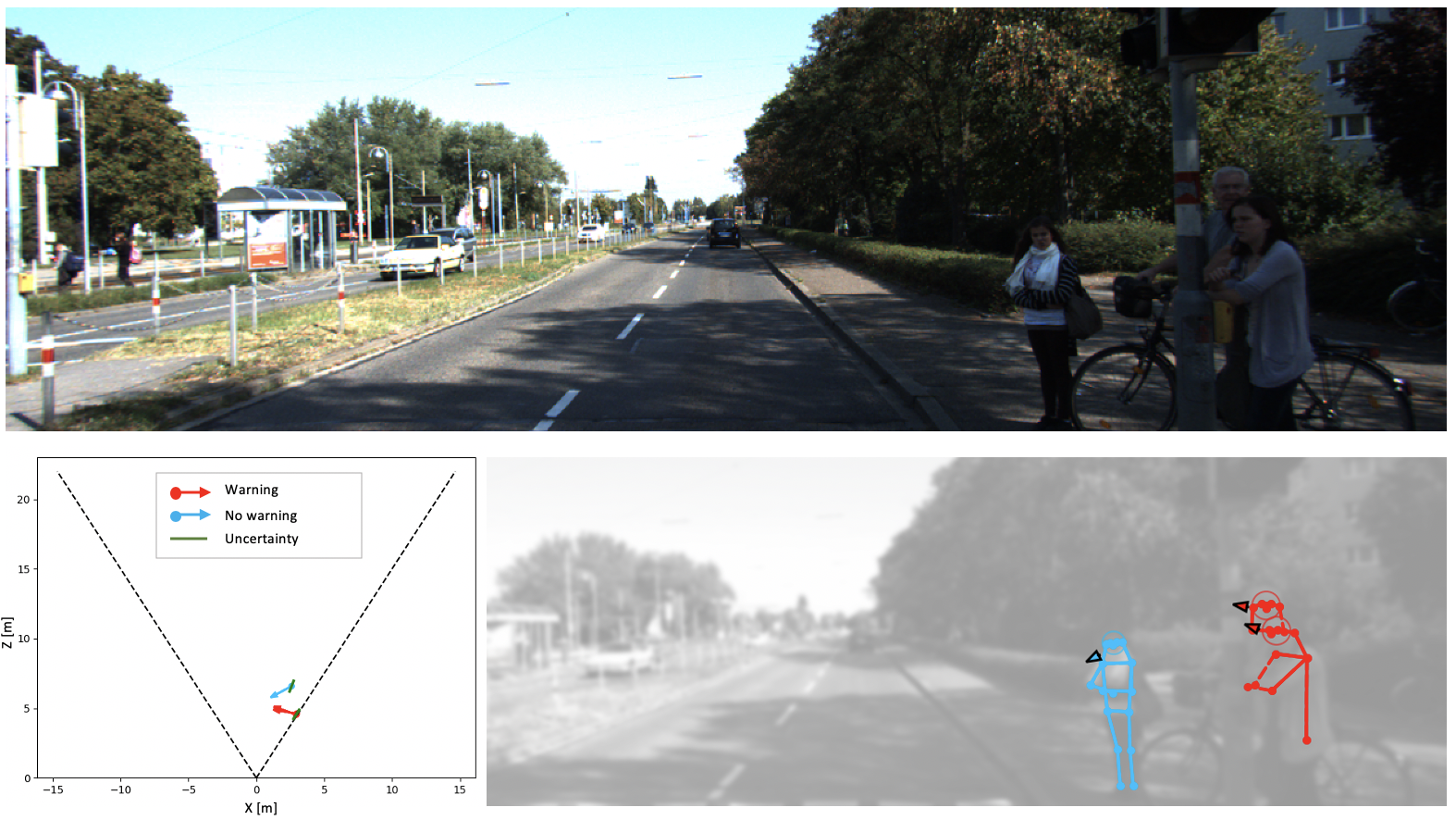}
    \caption{Qualitative results for the social distancing task in case of three people waiting at the traffic light. Two overlapping people are detected as very close to each other and the system warns for potential risk of contagion. A third person is located slightly more than two meters away and no warning is raised.}
  \label{fig:sd2}
\end{figure*}

\vspace{5pt}
\textbf{Results. }
Using the augmented KITTI dataset, we analyze whether social distancing is respected for 1760 people in the validation set. Using the ground-truth localization and orientation we generate labels for which 36.8\% of people do not comply with social distancing requirements. This is reasonable as the KITTI dataset contains many crowded scenes. As shown in Table \ref{tab:social_distance}, our MonoLoco++ obtains an accuracy of 84.0\%. We note that this dataset is more challenging than the Collective Activity one~\cite{choi2009they}, as it includes people 40+ meters away as well as occluded instances. Qualitative results are shown in Figures \ref{fig:sd1} and \ref{fig:sd2}, where our method estimates 3D localization and orientation, and verify social distancing compliance. In particular, Figure \ref{fig:sd2} shows that the network is able to accurately localize two overlapping people and recognize a potential risk of contagion, also based on people's relative orientation. In addition, we notice that orientation has a direct impact on reducing false alarms. Without orientation, the network estimates that 43 \% of instances violate social distancing requirements. Including orientation, the estimated number reaches 37\%, almost on par with the ground-truth value of 38\%.

\section{Privacy}
\begin{figure*}[t]
  \centering
    \includegraphics[width=0.95\linewidth]{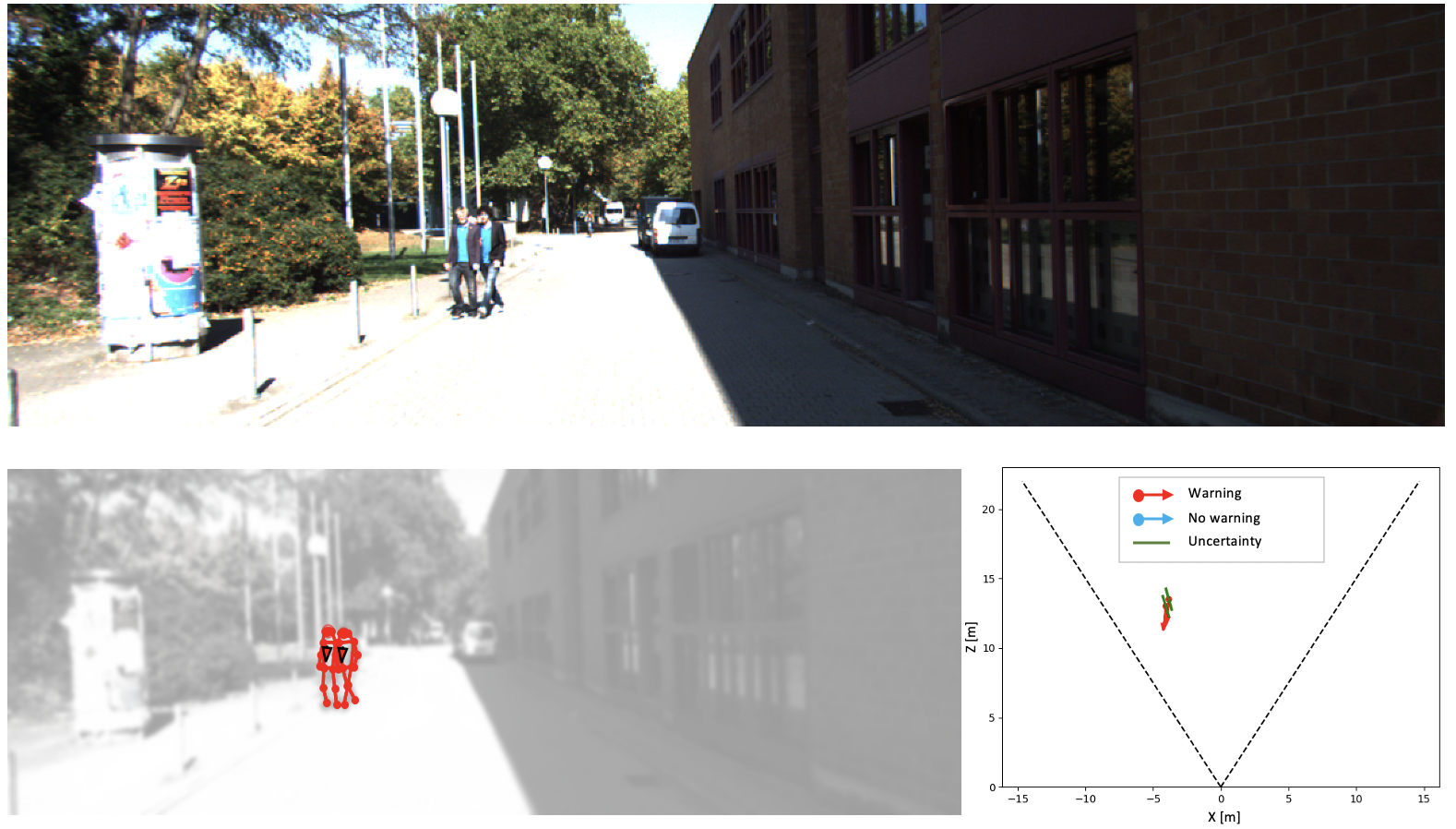}
    \caption{Qualitative results for the 3D localization task and social distancing. Our MonoLoco++ estimates 3D locations and orientations and raises a warning when social distancing is not respected.} 
  \label{fig:sd1}
\end{figure*}

Our network analyzes 2D poses and does not require any image to process the scene. In fact in Figures  \ref{fig:pull}, \ref{fig:sd1} and \ref{fig:sd2}, the original image is only shown to clarify the context, but it is not processed directly by MonoLoco++. We leverage an off-the-shelf pose detector which could be embedded in the camera itself. We have designed our system to encourage a privacy-by-design policy \cite{cristani2020visual}, where images are processed internally by smart cameras \cite{belbachir2010smart} and only 2D poses are sent remotely to a secondary system. The 2D poses do not contain any sensitive data but are informative enough to monitor social distancing. 

We also note that smart cameras differ from other technologies by being non-invasive and mostly non-collaborative \cite{cristani2020visual}. Differently from mobile applications, the user is not requested to share any personal data. On the contrary, a low-dimensional representation such as a 2D pose may be challenging for accurate 3D localization, but its ambiguity may prove useful for privacy concerns.

\section{Conclusions}

We have presented a new deep learning method that perceives humans' 3D locations and their body orientations from monocular cameras. We emphasized that the main challenge of perceiving social interactions is the ambiguity in 3D localizing people from a single image. Thus, we presented a method that predicts confidence intervals in contrast to point estimates leading to state-of-the-art results. Our system works with a single RGB image, shows cross-dataset generalization properties, and does not require homography calibration, making it suitable for fixed or mobile cameras already installed in transportation systems. 

While we have demonstrated the strengths of our method on popular tasks (monocular 3D localization and social interaction recognition), the COVID-19 outbreak has highlighted more than ever the need to perceive humans in 3D in the context of intelligent systems. We argued that to monitor social distancing effectively, we should go beyond a measure of distance. Orientations and relative positions of people strongly influence the risk of contagion, and people talking to each other incur higher risks than simply walking apart. Hence, we have presented an innovative approach to analyze social distancing, not only based on 3D localization  but also on social cues. We hope our work will also  contribute to the collective effort of preserving people's health while guaranteeing access to transportation hubs. 

\section{Acknowledgements}

This work is supported by the Swiss National Science Foundation under the Grant 2OOO21-L92326 and the SNSF Spark fund (190677). We also thank our lab members and reviewers for their valuable comments.

\FloatBarrier

\ifCLASSOPTIONcaptionsoff
  \newpage
\fi

\cleardoublepage
\bibliographystyle{IEEEtran}
\bibliography{references}

\begin{thebibliography}{100}
\providecommand{\url}[1]{#1}
\csname url@samestyle\endcsname
\providecommand{\newblock}{\relax}
\providecommand{\bibinfo}[2]{#2}
\providecommand{\BIBentrySTDinterwordspacing}{\spaceskip=0pt\relax}
\providecommand{\BIBentryALTinterwordstretchfactor}{4}
\providecommand{\BIBentryALTinterwordspacing}{\spaceskip=\fontdimen2\font plus
\BIBentryALTinterwordstretchfactor\fontdimen3\font minus
  \fontdimen4\font\relax}
\providecommand{\BIBforeignlanguage}[2]{{%
\expandafter\ifx\csname l@#1\endcsname\relax
\typeout{** WARNING: IEEEtran.bst: No hyphenation pattern has been}%
\typeout{** loaded for the language `#1'. Using the pattern for}%
\typeout{** the default language instead.}%
\else
\language=\csname l@#1\endcsname
\fi
#2}}
\providecommand{\BIBdecl}{\relax}
\BIBdecl

\bibitem{llorca2012stereo}
D.~Llorca, M.~Sotelo, A.~Hell{\'\i}n, A.~Orellana, M.~Gavil{\'a}n, I.~Daza, and
  A.~Lorente, ``Stereo regions-of-interest selection for pedestrian protection:
  A survey,'' \emph{Transportation research part C: emerging technologies},
  vol.~25, pp. 226--237, 2012.

\bibitem{alahi2011sparsity}
A.~Alahi, L.~Jacques, Y.~Boursier, and P.~Vandergheynst, ``Sparsity driven
  people localization with a heterogeneous network of cameras,'' \emph{Journal
  of Mathematical Imaging and Vision}, vol.~41, no. 1-2, pp. 39--58, 2011.

\bibitem{palffy2019occlusion}
A.~Palffy, J.~F. Kooij, and D.~M. Gavrila, ``Occlusion aware sensor fusion for
  early crossing pedestrian detection,'' in \emph{2019 IEEE Intelligent
  Vehicles Symposium (IV)}.\hskip 1em plus 0.5em minus 0.4em\relax IEEE, 2019,
  pp. 1768--1774.

\bibitem{arnold2019survey}
E.~Arnold, O.~Y. Al-Jarrah, M.~Dianati, S.~Fallah, D.~Oxtoby, and
  A.~Mouzakitis, ``A survey on 3d object detection methods for autonomous
  driving applications,'' \emph{IEEE Transactions on Intelligent Transportation
  Systems}, vol.~20, no.~10, pp. 3782--3795, 2019.

\bibitem{monoloco}
L.~Bertoni, S.~Kreiss, and A.~Alahi, ``Monoloco: Monocular 3d pedestrian
  localization and uncertainty estimation,'' in \emph{The IEEE International
  Conference on Computer Vision (ICCV)}, October 2019.

\bibitem{liu2020tanet}
Z.~Liu, X.~Zhao, T.~Huang, R.~Hu, Y.~Zhou, and X.~Bai, ``Tanet: Robust 3d
  object detection from point clouds with triple attention.'' in \emph{AAAI
  Conference on Artificial Intelligence}, 2020, pp. 11\,677--11\,684.

\bibitem{hotspotnet}
Q.~Chen, L.~Sun, Z.~Wang, K.~Jia, and A.~Yuille, ``Object as hotspots: An
  anchor-free 3d object detection approach via firing of hotspots,'' in
  \emph{The European Conference on Computer Vision (ECCV)}, 2020.

\bibitem{liang2018deep}
M.~Liang, B.~Yang, S.~Wang, and R.~Urtasun, ``Deep continuous fusion for
  multi-sensor 3d object detection,'' in \emph{The European Conference on
  Computer Vision (ECCV)}, 2018, pp. 641--656.

\bibitem{xu2018pointfusion}
D.~Xu, D.~Anguelov, and A.~Jain, ``Pointfusion: Deep sensor fusion for 3d
  bounding box estimation,'' in \emph{The IEEE Conference on Computer Vision
  and Pattern Recognition (CVPR)}, 2018, pp. 244--253.

\bibitem{alahi2014robust}
A.~Alahi, M.~Bierlaire, and P.~Vandergheynst, ``Robust real-time pedestrians
  detection in urban environments with low-resolution cameras,''
  \emph{Transportation research part C: emerging technologies}, vol.~39, pp.
  113--128, 2014.

\bibitem{delannay2009detection}
D.~Delannay, N.~Danhier, and C.~De~Vleeschouwer, ``Detection and recognition of
  sports (wo) men from multiple views,'' in \emph{2009 Third ACM/IEEE
  International Conference on Distributed Smart Cameras (ICDSC)}.\hskip 1em
  plus 0.5em minus 0.4em\relax IEEE, 2009, pp. 1--7.

\bibitem{hu20073d}
Z.~Hu, C.~Wang, and K.~Uchimura, ``3d vehicle extraction and tracking from
  multiple viewpoints for traffic monitoring by using probability fusion map,''
  in \emph{2007 IEEE Intelligent Transportation Systems Conference}.\hskip 1em
  plus 0.5em minus 0.4em\relax IEEE, 2007, pp. 30--35.

\bibitem{zhou2018voxelnet}
Y.~Zhou and O.~Tuzel, ``Voxelnet: End-to-end learning for point cloud based 3d
  object detection,'' in \emph{The IEEE Conference on Computer Vision and
  Pattern Recognition (CVPR)}, 2018, pp. 4490--4499.

\bibitem{qi2018frustum}
C.~R. Qi, W.~Liu, C.~Wu, H.~Su, and L.~J. Guibas, ``Frustum pointnets for 3d
  object detection from rgb-d data,'' in \emph{The IEEE Conference on Computer
  Vision and Pattern Recognition (CVPR)}, 2018, pp. 918--927.

\bibitem{cvd19worldbank}
W.~Bank, ``Protecting public transport from the coronavirus... and from
  financial collapse,'' 2020.

\bibitem{remington1985airborne}
P.~L. Remington, W.~N. Hall, I.~H. Davis, A.~Herald, and R.~A. Gunn, ``Airborne
  transmission of measles in a physician's office,'' \emph{Jama}, vol. 253,
  no.~11, pp. 1574--1577, 1985.

\bibitem{cvd19speaking}
E.~Bromage, ``The risks - know them - avoid them,'' 2020.

\bibitem{asadi2019aerosol}
S.~Asadi, A.~S. Wexler, C.~D. Cappa, S.~Barreda, N.~M. Bouvier, and W.~D.
  Ristenpart, ``Aerosol emission and superemission during human speech increase
  with voice loudness,'' \emph{Scientific reports}, vol.~9, no.~1, pp. 1--10,
  2019.

\bibitem{cvd19airborne}
V.~Stadnytskyi, C.~E. Bax, A.~Bax, and P.~Anfinrud, ``The airborne lifetime of
  small speech droplets and their potential importance in sars-cov-2
  transmission,'' \emph{Proceedings of the National Academy of Sciences}, 2020.

\bibitem{rummel1981understanding}
R.~J. Rummel, ``Understanding conflict and war: Vol. 5: The just peace,''
  \emph{Beverly Hills, California: Sage Publications}, 1981.

\bibitem{zhao2000mobile}
Y.~Zhao, ``Mobile phone location determination and its impact on intelligent
  transportation systems,'' \emph{IEEE Transactions on intelligent
  transportation systems}, vol.~1, no.~1, pp. 55--64, 2000.

\bibitem{zandbergen2009accuracy}
P.~A. Zandbergen, ``Accuracy of iphone locations: A comparison of assisted gps,
  wifi and cellular positioning,'' \emph{Transactions in GIS}, vol.~13, pp.
  5--25, 2009.

\bibitem{kasemsuppakorn2013pedestrian}
P.~Kasemsuppakorn and H.~A. Karimi, ``A pedestrian network construction
  algorithm based on multiple gps traces,'' \emph{Transportation research part
  C: emerging technologies}, vol.~26, pp. 285--300, 2013.

\bibitem{cristani2011social}
M.~Cristani, L.~Bazzani, G.~Paggetti, A.~Fossati, D.~Tosato, A.~Del~Bue,
  G.~Menegaz, and V.~Murino, ``Social interaction discovery by statistical
  analysis of f-formations.'' in \emph{British Machine Vision Conference
  (BMVC)}, vol.~2, 2011, p.~4.

\bibitem{cristani2011towards}
M.~Cristani, G.~Paggetti, A.~Vinciarelli, L.~Bazzani, G.~Menegaz, and
  V.~Murino, ``Towards computational proxemics: Inferring social relations from
  interpersonal distances,'' in \emph{2011 IEEE Third International Conference
  on Privacy, Security, Risk and Trust and 2011 IEEE Third International
  Conference on Social Computing}.\hskip 1em plus 0.5em minus 0.4em\relax IEEE,
  2011, pp. 290--297.

\bibitem{yang2012recognizing}
Y.~Yang, S.~Baker, A.~Kannan, and D.~Ramanan, ``Recognizing proxemics in
  personal photos,'' in \emph{the IEEE Conference on Computer Vision and
  Pattern Recognition (CVPR)}.\hskip 1em plus 0.5em minus 0.4em\relax IEEE,
  2012, pp. 3522--3529.

\bibitem{aimar2019social}
E.~S. Aimar, P.~Radeva, and M.~Dimiccoli, ``Social relation recognition in
  egocentric photostreams,'' in \emph{The IEEE International Conference on
  Image Processing (ICIP)}.\hskip 1em plus 0.5em minus 0.4em\relax IEEE, 2019,
  pp. 3227--3231.

\bibitem{kreiss2019pifpaf}
S.~Kreiss, L.~Bertoni, and A.~Alahi, ``Pifpaf: Composite fields for human pose
  estimation,'' in \emph{The IEEE Conference on Computer Vision and Pattern
  Recognition (CVPR)}, 2019, pp. 11\,977--11\,986.

\bibitem{kendon1990conducting}
A.~Kendon, \emph{Conducting interaction: Patterns of behavior in focused
  encounters}.\hskip 1em plus 0.5em minus 0.4em\relax CUP Archive, 1990,
  vol.~7.

\bibitem{Geiger2013Kitti}
A.~Geiger, P.~Lenz, C.~Stiller, and R.~Urtasun, ``Vision meets robotics: The
  kitti dataset,'' \emph{International Journal of Robotics Research (IJRR)},
  2013.

\bibitem{choi2009they}
W.~Choi, K.~Shahid, and S.~Savarese, ``What are they doing?: Collective
  activity classification using spatio-temporal relationship among people,'' in
  \emph{2009 IEEE 12th International Conference on Computer Vision Workshops,
  ICCV Workshops}.\hskip 1em plus 0.5em minus 0.4em\relax IEEE, 2009, pp.
  1282--1289.

\bibitem{lecun2015deep}
Y.~LeCun, Y.~Bengio, and G.~Hinton, ``Deep learning,'' \emph{nature}, vol. 521,
  no. 7553, pp. 436--444, 2015.

\bibitem{ren2015faster}
S.~Ren, K.~He, R.~Girshick, and J.~Sun, ``Faster r-cnn: Towards real-time
  object detection with region proposal networks,'' in \emph{Advances in neural
  information processing systems}, 2015, pp. 91--99.

\bibitem{yolo}
J.~Redmon, S.~Divvala, R.~Girshick, and A.~Farhadi, ``You only look once:
  Unified, real-time object detection,'' in \emph{the IEEE conference on
  computer vision and pattern recognition}, 2016, pp. 779--788.

\bibitem{openpose}
Z.~Cao, T.~Simon, S.-E. Wei, and Y.~Sheikh, ``Realtime multi-person 2d pose
  estimation using part affinity fields,'' in \emph{the IEEE Conference on
  Computer Vision and Pattern Recognition (CVPR)}, vol.~1, 2017, p.~7.

\bibitem{m3d}
X.~Chen, K.~Kundu, Z.~Zhang, H.~Ma, S.~Fidler, and R.~Urtasun, ``Monocular 3d
  object detection for autonomous driving,'' in \emph{The IEEE Conference on
  Computer Vision and Pattern Recognition (CVPR)}, 2016, pp. 2147--2156.

\bibitem{wenlongPSF}
W.~Deng, L.~Bertoni, S.~Kreiss, and A.~Alahi, ``Joint human pose estimation and
  stereo 3d localization,'' in \emph{the International Conference on Robotics
  and Automation (ICRA)}, 2020.

\bibitem{monstereo}
L.~Bertoni, S.~Kreiss, T.~Mordan, and A.~Alahi, ``Monstereo: When monocular and
  stereo meet at the tail of 3d human localization,'' in \emph{the
  International Conference on Robotics and Automation (ICRA)}, 2021.

\bibitem{muzahid2020curvenet}
A.~Muzahid, W.~Wan, F.~Sohel, L.~Wu, and L.~Hou, ``Curvenet: Curvature-based
  multitask learning deep networks for 3d object recognition,'' \emph{IEEE/CAA
  Journal of Automatica Sinica}, 2020.

\bibitem{godard2017monodepth}
C.~Godard, O.~Mac~Aodha, and G.~J. Brostow, ``Unsupervised monocular depth
  estimation with left-right consistency,'' in \emph{The IEEE Conference on
  Computer Vision and Pattern Recognition (CVPR)}, 2017, pp. 270--279.

\bibitem{alahi2017learning}
A.~Alahi, V.~Ramanathan, K.~Goel, A.~Robicquet, A.~A. Sadeghian, L.~Fei-Fei,
  and S.~Savarese, ``Learning to predict human behavior in crowded scenes,'' in
  \emph{Group and Crowd Behavior for Computer Vision}.\hskip 1em plus 0.5em
  minus 0.4em\relax Elsevier, 2017, pp. 183--207.

\bibitem{deng2009imagenet}
J.~Deng, W.~Dong, R.~Socher, L.-J. Li, K.~Li, and L.~Fei-Fei, ``Imagenet: A
  large-scale hierarchical image database,'' in \emph{2009 IEEE conference on
  computer vision and pattern recognition}.\hskip 1em plus 0.5em minus
  0.4em\relax Ieee, 2009, pp. 248--255.

\bibitem{Lin2014MicrosoftCC}
T.-Y. Lin, M.~Maire, S.~J. Belongie, L.~D. Bourdev, R.~B. Girshick, J.~Hays,
  P.~Perona, D.~Ramanan, P.~Doll{\'a}r, and C.~L. Zitnick, ``Microsoft coco:
  Common objects in context,'' in \emph{The European Conference on Computer
  Vision (ECCV)}, 2014.

\bibitem{sun2020proximity}
C.~{Sun}, J.~M.~U. {Vianney}, Y.~{Li}, L.~{Chen}, L.~{Li}, F.~{Wang},
  A.~{Khajepour}, and D.~{Cao}, ``Proximity based automatic data annotation for
  autonomous driving,'' \emph{IEEE/CAA Journal of Automatica Sinica}, vol.~7,
  no.~2, pp. 395--404, 2020.

\bibitem{nuscenes}
H.~Caesar, V.~Bankiti, A.~H. Lang, S.~Vora, V.~E. Liong, Q.~Xu, A.~Krishnan,
  Y.~Pan, G.~Baldan, and O.~Beijbom, ``nuscenes: A multimodal dataset for
  autonomous driving,'' \emph{arXiv preprint arXiv:1903.11027}, 2019.

\bibitem{argoverse}
M.-F. Chang, J.~W. Lambert, P.~Sangkloy, J.~Singh, S.~Bak, A.~Hartnett,
  D.~Wang, P.~Carr, S.~Lucey, D.~Ramanan, and J.~Hays, ``Argoverse: 3d tracking
  and forecasting with rich maps,'' in \emph{Conference on Computer Vision and
  Pattern Recognition (CVPR)}, 2019.

\bibitem{waymo}
P.~Sun, H.~Kretzschmar, X.~Dotiwalla, A.~Chouard, V.~Patnaik, P.~Tsui, J.~Guo,
  Y.~Zhou, Y.~Chai, B.~Caine \emph{et~al.}, ``Scalability in perception for
  autonomous driving: Waymo open dataset,'' in \emph{Proceedings of the
  IEEE/CVF Conference on Computer Vision and Pattern Recognition}, 2020, pp.
  2446--2454.

\bibitem{kothari_human_2020}
\BIBentryALTinterwordspacing
P.~Kothari, S.~Kreiss, and A.~Alahi, ``Human {Trajectory} {Forecasting} in
  {Crowds}: {A} {Deep} {Learning} {Perspective},'' \emph{arXiv:2007.03639
  [cs]}, Jul. 2020, arXiv: 2007.03639. [Online]. Available:
  \url{http://arxiv.org/abs/2007.03639}
\BIBentrySTDinterwordspacing

\bibitem{hall1966hidden}
E.~T. Hall, \emph{The hidden dimension}.\hskip 1em plus 0.5em minus 0.4em\relax
  Garden City, NY: Doubleday, 1966, vol. 609.

\bibitem{Papandreou2017TowardsAM}
G.~Papandreou, T.~Zhu, N.~Kanazawa, A.~Toshev, J.~Tompson, C.~Bregler, and
  K.~P. Murphy, ``Towards accurate multi-person pose estimation in the wild,''
  \emph{The IEEE Conference on Computer Vision and Pattern Recognition (CVPR)},
  pp. 3711--3719, 2017.

\bibitem{Fang2017RMPERM}
H.~Fang, S.~Xie, and C.~Lu, ``Rmpe: Regional multi-person pose estimation,''
  \emph{The IEEE International Conference on Computer Vision (ICCV)}, pp.
  2353--2362, 2017.

\bibitem{He2017MaskR}
K.~He, G.~Gkioxari, P.~Doll{\'a}r, and R.~B. Girshick, ``Mask r-cnn,''
  \emph{The IEEE International Conference on Computer Vision (ICCV)}, pp.
  2980--2988, 2017.

\bibitem{xiao2018simple}
B.~Xiao, H.~Wu, and Y.~Wei, ``Simple baselines for human pose estimation and
  tracking,'' in \emph{The European Conference on Computer Vision (ECCV)},
  2018, pp. 466--481.

\bibitem{Cao2017RealtimeM2}
Z.~Cao, T.~Simon, S.-E. Wei, and Y.~Sheikh, ``Realtime multi-person 2d pose
  estimation using part affinity fields,'' \emph{The IEEE Conference on
  Computer Vision and Pattern Recognition (CVPR)}, pp. 1302--1310, 2017.

\bibitem{newell2017associative}
A.~Newell, Z.~Huang, and J.~Deng, ``Associative embedding: End-to-end learning
  for joint detection and grouping,'' in \emph{Advances in Neural Information
  Processing Systems}, 2017, pp. 2277--2287.

\bibitem{personlab}
G.~Papandreou, T.~Zhu, L.-C. Chen, S.~Gidaris, J.~Tompson, and K.~Murphy,
  ``Personlab: Person pose estimation and instance segmentation with a
  bottom-up, part-based, geometric embedding model,'' in \emph{The European
  Conference on Computer Vision (ECCV)}, 2018, pp. 269--286.

\bibitem{kocabas2018multiposenet}
M.~Kocabas, S.~Karagoz, and E.~Akbas, ``Multiposenet: Fast multi-person pose
  estimation using pose residual network,'' in \emph{The European Conference on
  Computer Vision (ECCV)}, 2018, pp. 417--433.

\bibitem{kreiss2021openpifpaf}
S.~Kreiss, L.~Bertoni, and A.~Alahi, ``{OpenPifPaf: Composite Fields for
  Semantic Keypoint Detection and Spatio-Temporal Association},'' \emph{arXiv
  preprint arXiv:2103.02440}, March 2021.

\bibitem{martinez2017simple}
J.~Martinez, R.~Hossain, J.~Romero, and J.~J. Little, ``A simple yet effective
  baseline for 3d human pose estimation,'' in \emph{The IEEE International
  Conference on Computer Vision (ICCV)}.\hskip 1em plus 0.5em minus 0.4em\relax
  IEEE, 2017, pp. 2659--2668.

\bibitem{MorenoNoguer20173DHP}
F.~Moreno-Noguer, ``3d human pose estimation from a single image via distance
  matrix regression,'' \emph{The IEEE Conference on Computer Vision and Pattern
  Recognition (CVPR)}, pp. 1561--1570, 2017.

\bibitem{zanfir2018deep}
A.~Zanfir, E.~Marinoiu, M.~Zanfir, A.-I. Popa, and C.~Sminchisescu, ``Deep
  network for the integrated 3d sensing of multiple people in natural images,''
  in \emph{Advances in Neural Information Processing Systems}, 2018, pp.
  8410--8419.

\bibitem{Rogez2019LCRNetM2}
G.~Rogez, P.~Weinzaepfel, and C.~Schmid, ``Lcr-net++: Multi-person 2d and 3d
  pose detection in natural images,'' \emph{IEEE transactions on pattern
  analysis and machine intelligence}, 2019.

\bibitem{smpl2015}
M.~Loper, N.~Mahmood, J.~Romero, G.~Pons-Moll, and M.~J. Black, ``{SMPL}: A
  skinned multi-person linear model,'' \emph{ACM Trans. Graphics (Proc.
  SIGGRAPH Asia)}, vol.~34, no.~6, pp. 248:1--248:16, Oct. 2015.

\bibitem{kanazawa2018end}
A.~Kanazawa, M.~J. Black, D.~W. Jacobs, and J.~Malik, ``End-to-end recovery of
  human shape and pose,'' in \emph{the IEEE Conference on Computer Vision and
  Pattern Recognition (CVPR)}, 2018, pp. 7122--7131.

\bibitem{ku2019monopsr}
J.~Ku, A.~D. Pon, and S.~L. Waslander, ``Monocular 3d object detection
  leveraging accurate proposals and shape reconstruction,'' in \emph{The IEEE
  Conference on Computer Vision and Pattern Recognition (CVPR)}, 2019, pp.
  11\,867--11\,876.

\bibitem{monodis-tpami}
A.~Simonelli, S.~R. Bulo, L.~Porzi, M.~L. Antequera, and P.~Kontschieder,
  ``Disentangling monocular 3d object detection: From single to multi-class
  recognition,'' \emph{IEEE Transactions on Pattern Analysis and Machine
  Intelligence}, 2020.

\bibitem{smoke20}
Z.~Liu, Z.~Wu, and R.~Toth, ``Smoke: Single-stage monocular 3d object detection
  via keypoint estimation,'' in \emph{The IEEE Conference on Computer Vision
  and Pattern Recognition (CVPR) Workshops}, June 2020.

\bibitem{Kundegorski2014APA}
M.~E. Kundegorski and T.~P. Breckon, ``A photogrammetric approach for real-time
  3d localization and tracking of pedestrians in monocular infrared imagery,''
  in \emph{SPIE Optics and Photonics for Counterterrorism, Crime Fighting, and
  Defence}, vol. 9253, 2014.

\bibitem{alahi2015rgb}
A.~Alahi, A.~Haque, and L.~Fei-Fei, ``Rgb-w: When vision meets wireless,'' in
  \emph{The IEEE International Conference on Computer Vision (ICCV)}, 2015, pp.
  3289--3297.

\bibitem{alahi2008object}
A.~Alahi, M.~Bierlaire, and M.~Kunt, ``Object detection and matching with
  mobile cameras collaborating with fixed cameras,'' in \emph{Workshop on
  Multi-camera and Multi-modal Sensor Fusion Algorithms and Applications-M2SFA2
  2008}, 2008.

\bibitem{alahi2017tracking}
A.~Alahi, V.~Ramanathan, and L.~Fei-Fei, ``Tracking millions of humans in
  crowded spaces,'' in \emph{Group and Crowd Behavior for Computer
  Vision}.\hskip 1em plus 0.5em minus 0.4em\relax Elsevier, 2017, pp. 115--135.

\bibitem{mousavian20173d}
A.~Mousavian, D.~Anguelov, J.~Flynn, and J.~Kosecka, ``3d bounding box
  estimation using deep learning and geometry,'' in \emph{The IEEE Conference
  on Computer Vision and Pattern Recognition (CVPR)}, 2017, pp. 7074--7082.

\bibitem{qin2019monogrnet}
Z.~Qin, J.~Wang, and Y.~Lu, ``Monogrnet: A geometric reasoning network for
  monocular 3d object localization,'' in \emph{the AAAI Conference on
  Artificial Intelligence}, vol.~33, 2019, pp. 8851--8858.

\bibitem{Hu2018JointM3}
H.-N. Hu, Q.-Z. Cai, D.~Wang, J.~Lin, M.~Sun, P.~Krähenbühl, T.~Darrell, and
  F.~Yu, ``Joint monocular 3d vehicle detection and tracking,'' in \emph{The
  IEEE International Conference on Computer Vision (ICCV)}, 2019.

\bibitem{xu2018multi}
B.~Xu and Z.~Chen, ``Multi-level fusion based 3d object detection from
  monocular images,'' in \emph{The IEEE Conference on Computer Vision and
  Pattern Recognition (CVPR)}, 2018, pp. 2345--2353.

\bibitem{manhardt2019roi}
F.~Manhardt, W.~Kehl, and A.~Gaidon, ``Roi-10d: Monocular lifting of 2d
  detection to 6d pose and metric shape,'' in \emph{The IEEE Conference on
  Computer Vision and Pattern Recognition (CVPR)}, 2019, pp. 2069--2078.

\bibitem{roddick2018orthographic}
T.~Roddick, A.~Kendall, and R.~Cipolla, ``Orthographic feature transform for
  monocular 3d object detection,'' in \emph{the British Machine Vision
  Conference (BMVC)}, 2019.

\bibitem{Xiang2015Datadriven3V}
Y.~Xiang, W.~Choi, Y.~Lin, and S.~Savarese, ``Data-driven 3d voxel patterns for
  object category recognition,'' in \emph{The IEEE Conference on Computer
  Vision and Pattern Recognition (CVPR)}, 2015, pp. 1903--1911.

\bibitem{xiang2017subcategory}
------, ``Subcategory-aware convolutional neural networks for object proposals
  and detection,'' in \emph{The IEEE winter conference on applications of
  computer vision (WACV)}.\hskip 1em plus 0.5em minus 0.4em\relax IEEE, 2017,
  pp. 924--933.

\bibitem{Chabot2017DeepMA}
F.~Chabot, M.~A. Chaouch, J.~Rabarisoa, C.~Teuli{\`e}re, and T.~Chateau, ``Deep
  manta: A coarse-to-fine many-task network for joint 2d and 3d vehicle
  analysis from monocular image,'' in \emph{The IEEE Conference on Computer
  Vision and Pattern Recognition (CVPR))}, 2017, pp. 1827--1836.

\bibitem{3d-rcnn}
A.~Kundu, Y.~Li, and J.~M. Rehg, ``3d-rcnn: Instance-level 3d object
  reconstruction via render-and-compare,'' \emph{The IEEE Conference on
  Computer Vision and Pattern Recognition (CVPR)}, pp. 3559--3568, 2018.

\bibitem{Richard1991NeuralNC}
M.~D. Richard and R.~Lippmann, ``Neural network classifiers estimate bayesian a
  posteriori probabilities,'' \emph{Neural Computation}, vol.~3, pp. 461--483,
  1991.

\bibitem{neal2012bayesian}
R.~M. Neal, \emph{Bayesian learning for neural networks}.\hskip 1em plus 0.5em
  minus 0.4em\relax Springer Science \& Business Media, 2012, vol. 118.

\bibitem{graves2011practical}
A.~Graves, ``Practical variational inference for neural networks,'' in
  \emph{Advances in Neural Information Processing Systems}, 2011, pp.
  2348--2356.

\bibitem{blundell15weight}
C.~Blundell, J.~Cornebise, K.~Kavukcuoglu, and D.~Wierstra, ``Weight
  uncertainty in neural network,'' in \emph{the International Conference on
  Machine Learning}, ser. Proceedings of Machine Learning Research.\hskip 1em
  plus 0.5em minus 0.4em\relax PMLR, 2015, pp. 1613--1622.

\bibitem{salimans2015markov}
T.~Salimans, D.~Kingma, and M.~Welling, ``Markov chain monte carlo and
  variational inference: Bridging the gap,'' in \emph{the International
  Conference on Machine Learning}, 2015, pp. 1218--1226.

\bibitem{lakshminarayanan2017simple}
B.~Lakshminarayanan, A.~Pritzel, and C.~Blundell, ``Simple and scalable
  predictive uncertainty estimation using deep ensembles,'' in \emph{Advances
  in Neural Information Processing Systems}, 2017, pp. 6402--6413.

\bibitem{Gal2016Dropout}
Y.~Gal and Z.~Ghahramani, ``Dropout as a bayesian approximation: Representing
  model uncertainty in deep learning,'' in \emph{the International Conference
  on Machine Learning}, 2016, pp. 1050--1059.

\bibitem{gal2017concrete}
Y.~Gal, J.~Hron, and A.~Kendall, ``Concrete dropout,'' in \emph{Advances in
  Neural Information Processing Systems}, 2017, pp. 3581--3590.

\bibitem{srivastava2014dropout}
N.~Srivastava, G.~Hinton, A.~Krizhevsky, I.~Sutskever, and R.~Salakhutdinov,
  ``Dropout: a simple way to prevent neural networks from overfitting,''
  \emph{The Journal of Machine Learning Research}, vol.~15, no.~1, pp.
  1929--1958, 2014.

\bibitem{postels2019sampling}
J.~Postels, F.~Ferroni, H.~Coskun, N.~Navab, and F.~Tombari, ``Sampling-free
  epistemic uncertainty estimation using approximated variance propagation,''
  \emph{The IEEE International Conference of Computer Vision (ICCV)}, 2019.

\bibitem{Kendall2017WhatUD}
A.~Kendall and Y.~Gal, ``What uncertainties do we need in bayesian deep
  learning for computer vision?'' in \emph{Advances in Neural Information
  Processing Systems}, 2017, pp. 5574--5584.

\bibitem{mukhoti2018evaluating}
J.~Mukhoti and Y.~Gal, ``Evaluating bayesian deep learning methods for semantic
  segmentation,'' \emph{arXiv preprint arXiv:1811.12709}, 2018.

\bibitem{feng2018towards}
D.~Feng, L.~Rosenbaum, and K.~Dietmayer, ``Towards safe autonomous driving:
  Capture uncertainty in the deep neural network for lidar 3d vehicle
  detection,'' in \emph{the IEEE International Conference on Intelligent
  Transportation Systems (ITSC)}, 2018, pp. 3266--3273.

\bibitem{joo2019towards}
H.~Joo, T.~Simon, M.~Cikara, and Y.~Sheikh, ``Towards social artificial
  intelligence: Nonverbal social signal prediction in a triadic interaction,''
  in \emph{The IEEE Conference on Computer Vision and Pattern Recognition},
  2019, pp. 10\,873--10\,883.

\bibitem{kuzuoka2010reconfiguring}
H.~Kuzuoka, Y.~Suzuki, J.~Yamashita, and K.~Yamazaki, ``Reconfiguring spatial
  formation arrangement by robot body orientation,'' in \emph{2010 5th ACM/IEEE
  International Conference on Human-Robot Interaction (HRI)}.\hskip 1em plus
  0.5em minus 0.4em\relax IEEE, 2010, pp. 285--292.

\bibitem{tran2013social}
K.~N. Tran, A.~Bedagkar-Gala, I.~A. Kakadiaris, and S.~K. Shah, ``Social cues
  in group formation and local interactions for collective activity analysis.''
  in \emph{International Conference on Computer Vision Theory and Applications
  (VISAPP}, 2013, pp. 539--548.

\bibitem{bazzani2013social}
L.~Bazzani, M.~Cristani, D.~Tosato, M.~Farenzena, G.~Paggetti, G.~Menegaz, and
  V.~Murino, ``Social interactions by visual focus of attention in a
  three-dimensional environment,'' \emph{Expert Systems}, vol.~30, no.~2, pp.
  115--127, 2013.

\bibitem{vascon2014game}
S.~Vascon, E.~Z. Mequanint, M.~Cristani, H.~Hung, M.~Pelillo, and V.~Murino,
  ``A game-theoretic probabilistic approach for detecting conversational
  groups,'' in \emph{Asian Conference in Computer Vision (ACCV)}.\hskip 1em
  plus 0.5em minus 0.4em\relax Springer, 2014, pp. 658--675.

\bibitem{setti2015f}
F.~Setti, C.~Russell, C.~Bassetti, and M.~Cristani, ``F-formation detection:
  Individuating free-standing conversational groups in images,'' \emph{PloS
  one}, vol.~10, no.~5, p. e0123783, 2015.

\bibitem{aghaei2015towards}
M.~Aghaei, M.~Dimiccoli, and P.~Radeva, ``Towards social interaction detection
  in egocentric photo-streams,'' in \emph{Eighth International Conference on
  Machine Vision (ICMV 2015)}, vol. 9875.\hskip 1em plus 0.5em minus
  0.4em\relax International Society for Optics and Photonics, 2015, p. 987514.

\bibitem{nakamura2017jointly}
K.~Nakamura, S.~Yeung, A.~Alahi, and L.~Fei-Fei, ``Jointly learning energy
  expenditures and activities using egocentric multimodal signals,'' in
  \emph{Proceedings of the IEEE Conference on Computer Vision and Pattern
  Recognition}, 2017, pp. 1868--1877.

\bibitem{ioffe2015batch}
S.~Ioffe and C.~Szegedy, ``Batch normalization: Accelerating deep network
  training by reducing internal covariate shift,'' \emph{arXiv preprint
  arXiv:1502.03167}, 2015.

\bibitem{se2002ground}
S.~Se and M.~Brady, ``Ground plane estimation, error analysis and
  applications,'' \emph{Robotics and Autonomous systems}, vol.~39, no.~2, pp.
  59--71, 2002.

\bibitem{3dop}
X.~Chen, K.~Kundu, Y.~Zhu, A.~G. Berneshawi, H.~Ma, S.~Fidler, and R.~Urtasun,
  ``3d object proposals for accurate object class detection,'' in
  \emph{Advances in Neural Information Processing Systems}, 2015, pp. 424--432.

\bibitem{visscher2008sizing}
P.~M. Visscher, ``Sizing up human height variation,'' \emph{Nature genetics},
  vol.~40, no.~5, p. 489, 2008.

\bibitem{freeman1995cross}
J.~Freeman, T.~Cole, S.~Chinn, P.~Jones, E.~White, and M.~Preece, ``Cross
  sectional stature and weight reference curves for the uk, 1990.''
  \emph{Archives of disease in childhood}, vol.~73, no.~1, pp. 17--24, 1995.

\bibitem{li2019stereo}
P.~Li, X.~Chen, and S.~Shen, ``Stereo r-cnn based 3d object detection for
  autonomous driving,'' in \emph{The IEEE Conference on Computer Vision and
  Pattern Recognition (CVPR)}, 2019, pp. 7644--7652.

\bibitem{he2016residual}
K.~He, X.~Zhang, S.~Ren, and J.~Sun, ``Deep residual learning for image
  recognition,'' in \emph{The IEEE Conference on Computer Vision and Pattern
  Recognition}, 2016, pp. 770--778.

\bibitem{der2009aleatory}
A.~Der~Kiureghian and O.~Ditlevsen, ``Aleatory or epistemic? does it matter?''
  \emph{Structural Safety}, vol.~31, no.~2, pp. 105--112, 2009.

\bibitem{wirges2019capturing}
S.~Wirges, M.~Reith-Braun, M.~Lauer, and C.~Stiller, ``Capturing object
  detection uncertainty in multi-layer grid maps,'' \emph{arXiv preprint
  arXiv:1901.11284}, 2019.

\bibitem{kingma2014adam}
D.~P. Kingma and J.~Ba, ``Adam: A method for stochastic optimization,''
  \emph{arXiv preprint arXiv:1412.6980}, 2014.

\bibitem{pytorch}
A.~Paszke, S.~Gross, F.~Massa, A.~Lerer, J.~Bradbury, G.~Chanan, T.~Killeen,
  Z.~Lin, N.~Gimelshein, L.~Antiga \emph{et~al.}, ``Pytorch: An imperative
  style, high-performance deep learning library,'' in \emph{Advances in Neural
  Information Processing Systems}, 2019, pp. 8024--8035.

\bibitem{drillis1969body}
R.~Drillis, R.~Contini, and M.~Bluestein, \emph{Body segment parameters}.\hskip
  1em plus 0.5em minus 0.4em\relax New York University, School of Engineering
  and Science, 1969.

\bibitem{virtualsamples19}
H.~{Han}, M.~{Zhou}, and Y.~{Zhang}, ``Can virtual samples solve small sample
  size problem of kissme in pedestrian re-identification of smart
  transportation?'' \emph{IEEE Transactions on Intelligent Transportation
  Systems}, vol.~21, no.~9, pp. 3766--3776, 2020.

\bibitem{caetano2019skeleton}
C.~Caetano, F.~Br{\'e}mond, and W.~R. Schwartz, ``Skeleton image representation
  for 3d action recognition based on tree structure and reference joints,'' in
  \emph{2019 32nd SIBGRAPI Conference on Graphics, Patterns and Images
  (SIBGRAPI)}.\hskip 1em plus 0.5em minus 0.4em\relax IEEE, 2019, pp. 16--23.

\bibitem{bagautdinov2017social}
T.~Bagautdinov, A.~Alahi, F.~Fleuret, P.~Fua, and S.~Savarese, ``Social scene
  understanding: End-to-end multi-person action localization and collective
  activity recognition,'' in \emph{Proceedings of the IEEE conference on
  computer vision and pattern recognition}, 2017, pp. 4315--4324.

\bibitem{gavrilyuk2020actor}
K.~Gavrilyuk, R.~Sanford, M.~Javan, and C.~G. Snoek, ``Actor-transformers for
  group activity recognition,'' in \emph{the IEEE/CVF Conference on Computer
  Vision and Pattern Recognition}, 2020, pp. 839--848.

\bibitem{cristani2020visual}
M.~{Cristani}, A.~D. {Bue}, V.~{Murino}, F.~{Setti}, and A.~{Vinciarelli},
  ``The visual social distancing problem,'' \emph{IEEE Access}, vol.~8, pp.
  126\,876--126\,886, 2020.

\bibitem{belbachir2010smart}
A.~N. Belbachir, \emph{Smart cameras}.\hskip 1em plus 0.5em minus 0.4em\relax
  Springer, 2010, vol.~2.

\end{thebibliography}
\begin{IEEEbiography}[{\includegraphics[width=1.05in,height=1.5in,clip,keepaspectratio]{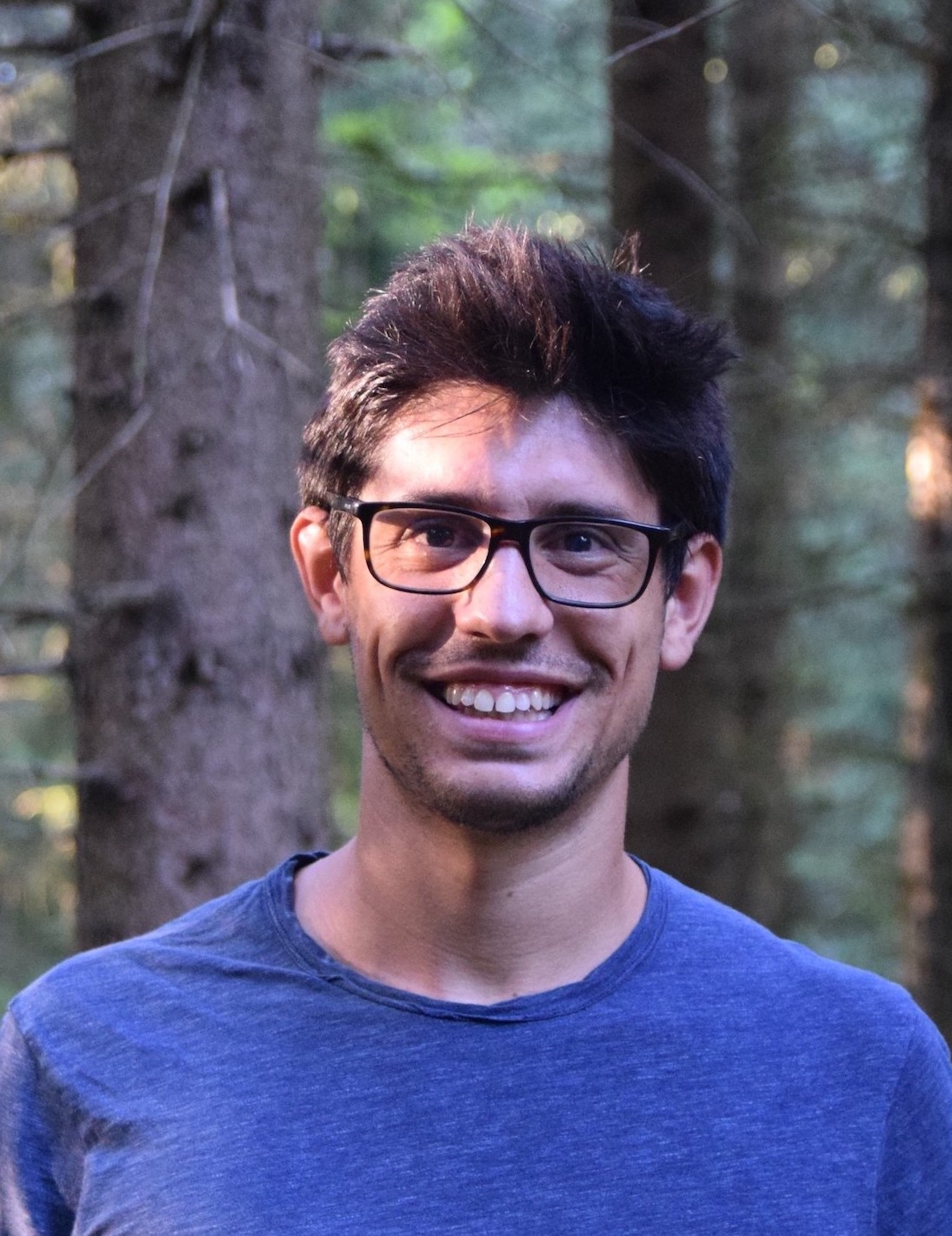}}]{Lorenzo Bertoni} is a doctoral student at the Visual Intelligence for Transportation (VITA) lab at EPFL in Switzerland focusing on 3D vision for vulnerable road users. Before joining EPFL, Lorenzo was a management consultant at Oliver Wyman and a visiting researcher at the University of California, Berkeley, working on predictive control for autonomous vehicles. Lorenzo received Bachelors and Masters Degrees in Engineering from the Polytechnic University of Turin and the University of Illinois at Chicago.  
\end{IEEEbiography}

\begin{IEEEbiography}[{\includegraphics[width=1.05
in,height=1.5in,clip,keepaspectratio]{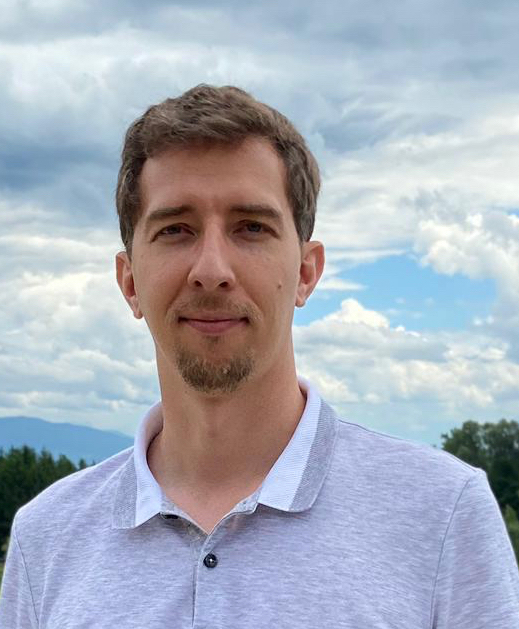}}]{Sven Kreiss} is a postdoc at the Visual Intelligence for Transportation (VITA) lab at EPFL in Switzerland focusing on perception with composite fields. Before returning to academia, he was the Senior Data Scientist at Sidewalk Labs~(Alphabet, Google sister) and worked on geospatial machine learning for urban environments. Prior to his industry experience, Sven developed statistical tools and methods used in particle physics research.
\end{IEEEbiography} 

\begin{IEEEbiography}[{\includegraphics[width=1.05in,height=1.25in,clip,keepaspectratio]{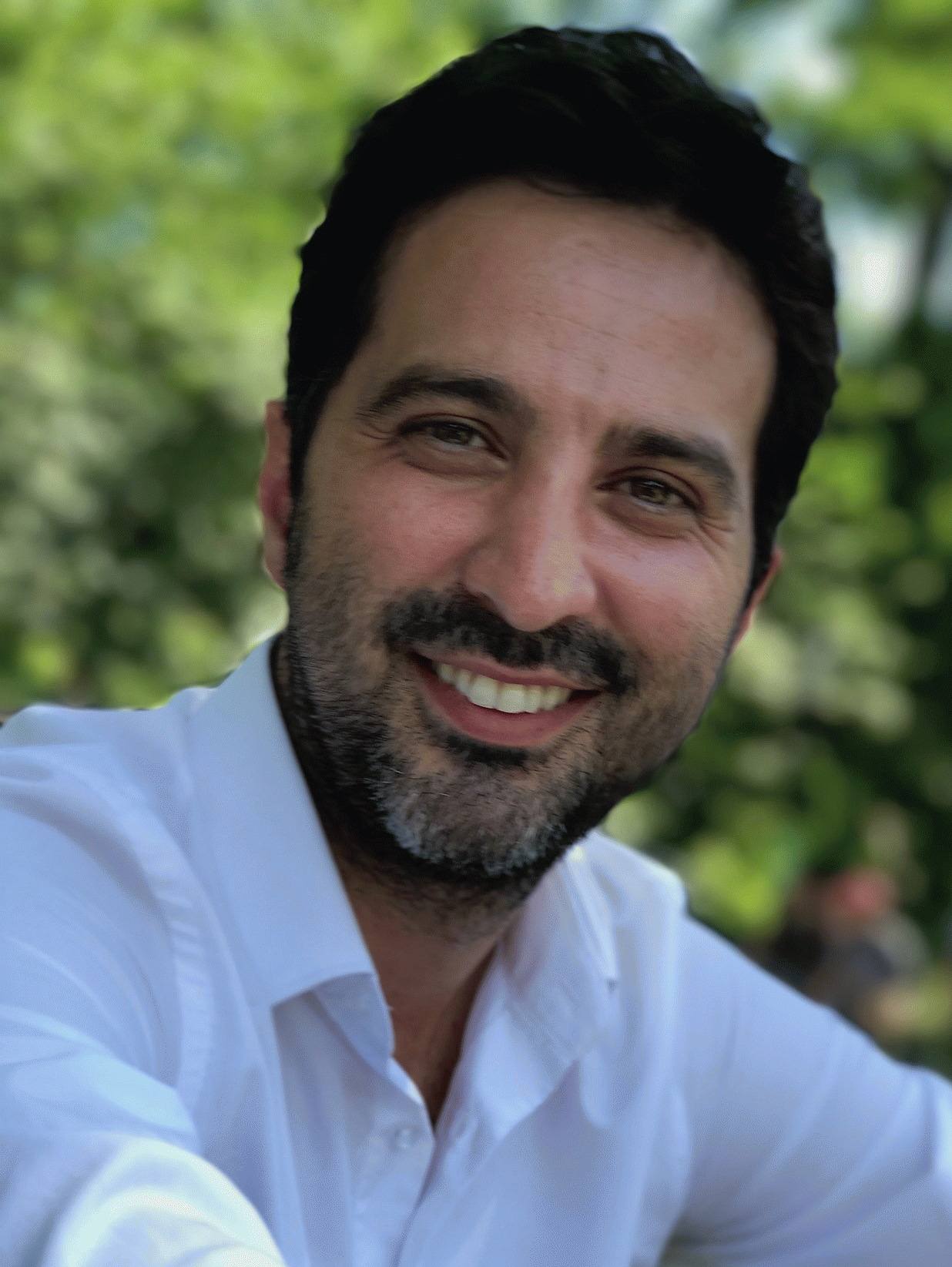}}]{Alexandre Alahi} is an Assistant Professor at EPFL. He spent five years at Stanford University as a Post-doc and Research Scientist after obtaining his Ph.D. from EPFL. His research enables machines to perceive the world and make decisions in the context of transportation problems and smart environments. He has worked on the theoretical challenges and practical applications of socially-aware Artificial Intelligence, i.e., systems equipped with perception and social intelligence. He was awarded the Swiss NSF early and advanced researcher grants for his work on predicting human social behavior. 
Alexandre has also co-founded multiple startups such as Visiosafe, and won several startup competitions. 
He was elected as one of the Top 20 Swiss Venture leaders in 2010.
\end{IEEEbiography}

\vfill

\end{document}